%% file: bare_jrnl_compsoc.tex
\documentclass[10pt,journal,compsoc]{IEEEtran}
\usepackage{graphicx}
\usepackage{amsmath}
\usepackage{amssymb}
\usepackage{booktabs}

\usepackage[utf8]{inputenc} 
\usepackage[T1]{fontenc}    
\usepackage{url}            
\usepackage{booktabs}       
\usepackage{amsfonts}       
\usepackage{nicefrac}       
\usepackage{microtype}      
\usepackage{xcolor}         
\usepackage{soul}

\usepackage{unicode}
\usepackage{orcidlink}

\usepackage{epsfig}
\usepackage{graphicx}
\usepackage{amsmath}
\usepackage{amssymb}
\usepackage{multirow}
\usepackage{xcolor}
\usepackage{algorithmic}
\usepackage{algorithm}
\usepackage{booktabs}
\usepackage{url}

\usepackage{colortbl}
\usepackage{bm}
\usepackage{subcaption}
\usepackage{makecell}
\usepackage{pifont}
\usepackage{enumitem}
\usepackage{wrapfig}

\definecolor{gain}{rgb}{0.224, 0.710, 0.290}

\newcommand{\xmark}{\ding{55}}
\newcommand{\cmark}{\ding{51}}%

\newcommand{\del}[1]{}

\newcommand{\etal}[1]{\textit{et al.}}

\ifCLASSOPTIONcompsoc
  \usepackage[nocompress]{cite}
\else
  \usepackage{cite}
\fi
\ifCLASSINFOpdf
\else
\fi
\begin{document}
%
\title{SGTR+: End-to-end Scene Graph Generation with Transformer}
%
%
%
%

\author{Rongjie~Li$^{\orcidlink{0000-0003-0675-9504}}$ ,
        Songyang~Zhang$^{\orcidlink{0000-0002-1902-1720}}$ ,
        and~Xuming~He$^{\orcidlink{0000-0003-2150-1237}}$ 
\IEEEcompsocitemizethanks{
\IEEEcompsocthanksitem Rongjie Li and Songyang Zhang are with ShanghaiTech University, Shanghai 201210, China (e-mail: lirj2@shanghaitech.edu.cn; zhangsongyang@pjlab.org.cn).
\IEEEcompsocthanksitem Xuming He is with the ShanghaiTech University, Shanghai Engineering Research Center of Intelligent Vision and Imaging, Pudong, Shanghai 201210, China. (e-mail: hexm@shanghaitech.edu.cn).
\IEEEcompsocthanksitem The corresponding author is Xuming He.
\IEEEcompsocthanksitem This work was supported by Shanghai Science and Technology Program 21010502700, Shanghai Frontiers Science Center of Human-centered Artificial Intelligence, and the National Nature Science Foundation of China under Grant 62350610269.
}}

%
%

\markboth{Journal of \LaTeX\ Class Files,~Vol.~14, No.~8, August~2015}%
{Shell \MakeLowercase{\textit{et al.}}: Bare Demo of IEEEtran.cls for Computer Society Journals}
%



\IEEEtitleabstractindextext{
\begin{abstract}
  Scene Graph Generation (SGG) remains a challenging visual understanding task due to its compositional property. Most previous works adopt a bottom-up, two-stage or point-based, one-stage approach, which often suffers from high time complexity or suboptimal designs.
  In this work, we propose a novel SGG method to address the aforementioned issues, formulating the task as a bipartite graph construction problem. To address the issues above, we create a transformer-based end-to-end framework to generate the entity and entity-aware predicate proposal set, and infer directed edges to form relation triplets.
  Moreover, we design a graph assembling module to infer the connectivity of the bipartite scene graph based on our entity-aware structure, enabling us to generate the scene graph in an end-to-end manner.
  Based on bipartite graph assembling paradigm, we further propose a new technical design to address the efficacy of entity-aware modeling and optimization stability of graph assembling.
  Equipped with the enhanced entity-aware design, our method achieves optimal performance and time-complexity.
  Extensive experimental results show that our design is able to achieve the state-of-the-art or comparable performance on three challenging benchmarks, surpassing most of the existing approaches and enjoying higher efficiency in inference. 
  Code is available: \url{https://github.com/Scarecrow0/SGTR}

\end{abstract}

\begin{IEEEkeywords}
  Computer Vision, Deep Learning,  Scene Graph Generation, Scene Understanding, Visual Relationship Detection
\end{IEEEkeywords}}

\maketitle
\IEEEdisplaynontitleabstractindextext

%
\IEEEpeerreviewmaketitle

\input{sec/intro.tex}

\input{sec/related}

\input{sec/method}

\input{sec/exp}

\input{sec/conclu}

\ifCLASSOPTIONcaptionsoff
  \newpage
\fi

\newpage
\appendices
\input{sec/supply}



%
{	\small
	\bibliographystyle{IEEEtran}
	\bibliography{main}
}

%




\begin{IEEEbiography}[{\includegraphics[width=1in,height=1.25in,clip,keepaspectratio]{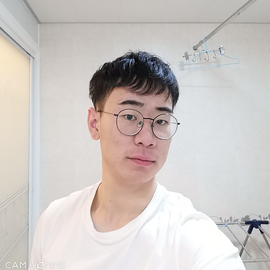}}]{Rongjie Li}
received the B.E. degree in the school of computer science and engineering from Northeastern University, Shenyang, China, in 2019. He is currently pursuing the Ph.D degree in computer science and technology at the ShanghaiTech University, supervised Prof. Xuming He. His research interests concern computer vision and machine learning.
\end{IEEEbiography}

\begin{IEEEbiography}[{\includegraphics[width=1in,height=1.25in,clip,keepaspectratio]{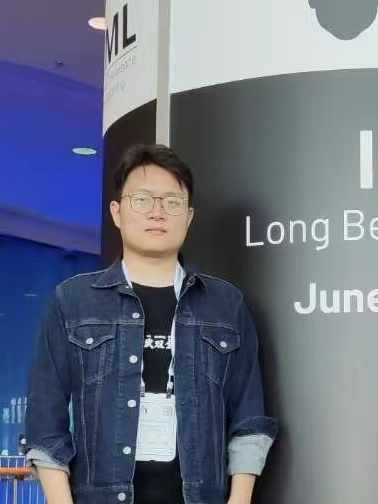}}]{Songyang Zhang}
received the Ph.D. degree in Computer Science at the University of Chinese Academy of Science, in the joint program at PLUS Lab, ShanghaiTech University, supervised Prof. Xuming He in 2022. He is currently work as Postdoctoral Fellow at Shanghai AI Laboratory, Shanghai, China. His research interests concern computer vision and machine learning.
\end{IEEEbiography}


\begin{IEEEbiography}[{\includegraphics[width=1in,height=1.25in,clip,keepaspectratio]{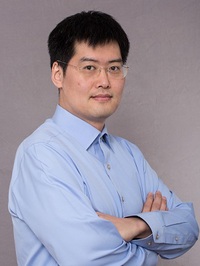}}]{Xuming He}
received the Ph.D. degree in computer science from the University of Toronto, Toronto, ON, Canada, in 2008. He held a Post-Doctoral position with the University of California at Los Angeles, Los Angeles, CA, USA, from 2008 to 2010. He was an Adjunct Research Fellow with The Australian National University, Canberra, ACT, Australia, from 2010 to 2016. He joined National ICT Australia, Canberra, where he was a Senior Researcher from 2013 to 2016. He is currently an Associate Professor with the School of Information Science and Technology, ShanghaiTech University, Shanghai, China. His research interests include semantic image and video segmentation, 3-D scene understanding, visual motion analysis, and efficient inference and learning in structured models.
\end{IEEEbiography}

\end{document}

%% file: sec/intro.tex
\section{Introduction}\label{sec:intro}

\IEEEPARstart{I}{nferring} structural properties of a scene, such as the relationships between visual entities, is a fundamental scene understanding task. 
A visual relationship between two entities typically consists of a triple \textit{ <subject entity, predicate, object entity>}. Given such relationships, a scene can be represented as a graph structure with the entities as nodes and the predicates as edges, which is referred to as a scene graph.
The scene graph provides a compact structural representation for a visual scene with potential applications in many vision tasks such as visual question answering~\cite{teney2017graph, shi2019explainable, hildebrandt2020scene, chang2021comprehensive}, image captioning~\cite{yang2019auto, yang2021reformer} and image retrieval~\cite{johnson2015image}.


Different from the traditional vision tasks focusing on visual entities (\textit{e.g.,} object detection), the problem of scene graph generation (SGG) requires jointly modeling the subject/object entities and their relations. 
The compositional property of visual relationships induces large variations in their appearance and relatively complex prediction space. Therefore, it is challenging to learn an efficient model to effectively localize and classify a variety of relationship concepts.

\begin{figure}[h!]
    \centering
    \includegraphics[width=0.96\linewidth]{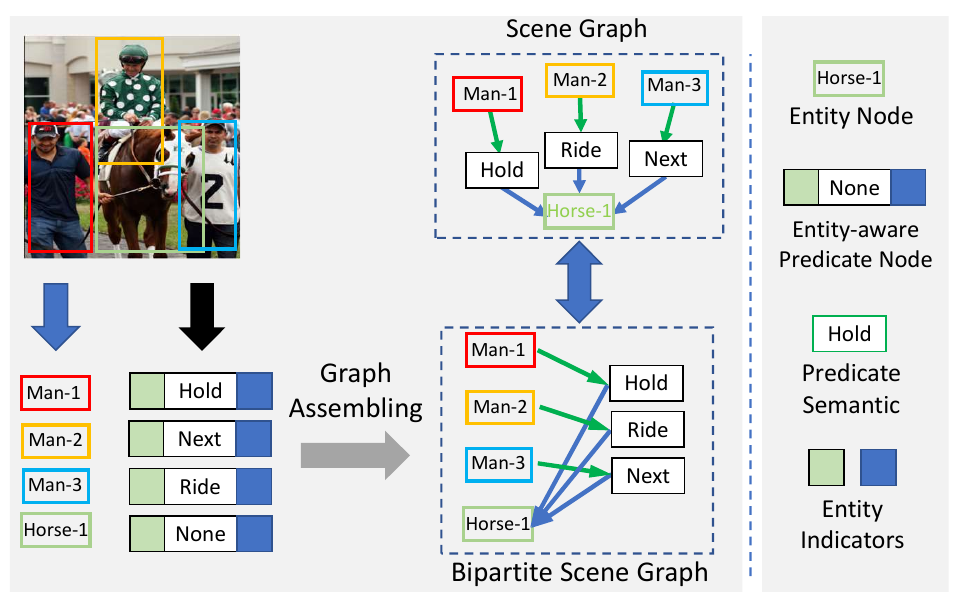}
    \vspace{-0.1cm}
    \caption{\textbf{An overview of SGTR pipeline for scene graph generation.}
    We formulate SGG as a bipartite graph construction process. First, the entity and predicate nodes are generated, respectively.
    Then we assemble the bipartite scene graph from two types of nodes.
     }
     \vspace{-0.4cm}
    \label{fig:idea}
\end{figure}

Most previous works attempt to tackle this problem using two distinct design patterns: \textit{bottom-up two-stage} methods~\cite{li2021bipartite,yang2021probabilistic, yao2021visual, desai2021learning, chiou2021recovering, guo2021general, knyazev2021generative, abdelkarim2021exploring} and \textit{point-based one-stage} methods~\cite{liu2021fully, dong2021visual}. 
The former typically first detects $N$ entity proposals, followed by predicting the predicate categories of those entity combinations. While this strategy achieves high recalls in discovering relation instances, its $\mathcal{O}(N^2)$ predicate proposals not only incur considerable computation cost but also produce substantial noise in context modeling.
The one-stage methods, on the other hand, extract entities and predicates separately from the images in order to reduce the size of relation proposal set.  
Nonetheless, they rely on a strong assumption of the non-overlapping property of interaction regions, which severely restricts their application in modeling complex scenes\footnote{
Such limitations, \textit{e.g.,} two different relationships cannot share a similar image region, are also discussed in the related literature~\cite{chen2021reformulating, tamura2021qpic}.}.

In this work, we aim to tackle the aforementioned limitation by leveraging the compositional property of scene graphs. 
To this end, as illustrated in Fig.~\ref{fig:idea}, we first formulate the SGG task as a bipartite graph construction problem, in which each relationship triplet is represented as two types of nodes (entity and predicate) linked by directed edges. 
Such a bipartite graph allows us to jointly generate entity/predicate nodes and their potential associations, yielding a rich hypothesis space for inferring visual relations. 
More importantly, we propose a novel entity-aware predicate representation that incorporates relevant entity proposal information into each predicate node. 
This enriches the predicate representations and therefore enables us to produce a relatively small number of high-quality predicate proposals. 
Moreover, such a representation encodes potential associations between each predicate and its subject/object entities, which can facilitate predicting the graph edges and lead to efficient generation of the visual relation triplets.

Specifically, we develop a new transformer-based end-to-end SGG model, dubbed Scene graph Generation TRansformer (SGTR), for constructing the bipartite graph. Our model consists of three main modules, including an \textit{entity node generator}, a \textit{predicate node generator} and a \textit{graph assembling module}.
Given an image, we first adopt a CNN+Transformer encoder to extract a set of image features. Those features are fed into two Transformer decoder networks, referred to as entity and predicate node generator, to produce a set of entity and predicate proposals, respectively. 
Our entity node generator uses a DETR-like decoder to detect a set of visual objects while the predicate node generator uses three parallel transformer decoders to compute a set of \textit{structured predicate representations}, each of which is then decoded into a predicate with its entity indicators. These indicators represent the subject/object entities associated with the predicate and hence we refer to the predicate generator's output as the entity-aware predicate proposals. 
After generating entity and predicate nodes (i.e., proposals), we then devise a differentiable \textit{graph assembling} module to infer the directed edges of the bipartite graph, which exploits the entity-aware predicate representation to predict the best grouping of the entity and predicate nodes.
With end-to-end training, our SGTR learns to infer a sparse set of relationship triplets from both the input image and entity proposals, which mitigates the impact of noisy object detection.


{
The original design of the SGTR, as presented in the conference version \cite{li2022sgtr} of this work, predominantly relies on entity features for computing the predicate-entity associations.  
It's worth noting that this representation tuned for entity estimation tends to be less contextual compared to the initial image features, potentially impacting the quality of entity-aware predicate nodes\footnote{The high-quality entity-aware representations play a critical role in entity-predicate matching during graph assembly, as demonstrated in the experimental section (refer to supplementary)}.
Additionally, its multiple entity-related decoding from the full image could potentially introduce additional computational overhead. 
Furthermore, the graph assembly process employs a heuristic distance function and a top-k selection strategy for linking edges. Unfortunately, this approach lacks the ability to be jointly optimized alongside node generators, thereby hindering its capacity to handle diverse entity-predicate pairings.
}

{
In order to address these challenges, we present SGTR+, an enhanced technical design of the SGTR that introduces two significant improvements. 
Firstly, we introduce a novel \textit{spatial-aware predicate node generator} that utilizes explicit spatial cues from entity nodes. This enables us to decode entity-aware predicates directly from image features, eliminating the need for the second decoding from the entire image.
This simple yet effective design enhances the efficiency and quality of entity-aware predicate nodes.
Secondly, we introduce a parameterized \textit{unified graph assembly} module. This module incorporates a learnable entity-predicate association embedding, facilitating fully differentiable graph assembly and enabling joint optimization with node generators. As a result, our enhanced design adeptly copes with a wide range of entity qualities during the graph assembly process.
}

We validate our method by extensive experiments on three public SGG datasets, including Visual Genome~\cite{krishna2017visual}, OpenImages-V6~\cite{OpenImages} and GQA~\cite{hudson2018gqa}. Our results show that the newly-designed SGTR+ outperforms previous state-of-the-art methods by a considerable margin in the mean recall metric and achieves higher efficiency in inference and model complexity.

The main contribution of our work has four-folds: 
\begin{itemize}[itemsep=0mm,topsep=0pt]
    \item We propose a novel transformer-based scene graph generation method with a bipartite graph construction process that inherits the advantages of both two-stage and one-stage methods.
    \item We develop an entity-aware predicate representation for exploiting the compositional properties of visual relationships.
    \item We further enhance entity-predicate association modeling by redesigning the main module of the predicate node generator and graph assembly, based on the SGTR architecture from the conference version.
    \item Our method achieves the state-of-the-art or comparable performance on all metrics w.r.t the prior SGG methods and with more efficient inference. 
\end{itemize}

%% file: sec/related.tex
\vspace{-0.1cm}
\section{Related Work}
\vspace{-0.1cm}

The problem of inferring visual relations from images has attracted much interest in computer vision, and there have been a large body of literature on general SGG as well as its related task on Human-Object Interaction (HOI). Here we categorize the existing approaches of SGG and HOI into three groups according to their task and strategy, with emphasis on the one-stage methods. 

\noindent\textbf{Two-stage Scene Graph Generation}
Two-stage SGG methods typically first generate relationship proposals using densely-connected entity pairs, followed by classifying the proposals with predicate classes. One major research effort in two-stage SGG, including many early works, is to better capture contextual information in the entity and relation proposals. To this end, a large variety of graph-based models with different structures has been explored for context modeling, which include 
sequential models~\cite{zellers_neural_2017}, fully-connected graphs~\cite{ xu_scene_2017, li_scene_2017, woo_linknet:_2018, li_factorizable_2018, yin_zoom-net:_2018, wang_exploring_2019, lin_gps-net_2020, cong_nodis_nodate, wang2020tackling, khandelwal2021segmentation}, dynamic/sparse graphs~\cite{tang_learning_2018, qi_attentive_2018,yang_graph_2018, li2021bipartite, lin2022ru} and graphs augmented by external knowledge~\cite{ zareian_bridging_2020, zareian2020weakly,zareian_learning_2020}.
More recent studies focus on the long-tail annotation issue in typical benchmarks and develop a series of balancing strategies to reduce the influence of language bias and label imbalance, such as inference time logit adjustment~\cite{tang_unbiased_2020, suhail2021energy, yan_pcpl_2020, chiou2021recovering, guo2021general, sun2023unbiased, liu2023neural} and training time loss re-weighting~\cite{yang2021probabilistic, knyazev_graph_2020, wang2020tackling, li2022ppdl}, or data resampling~\cite{li2021bipartite, yao2021visual, desai2021learning, knyazev2021generative, abdelkarim2021exploring, li2022devil, dong2022stacked}.

The two-stage methods have achieved strong performance in SGG as they are able to explore a large hypothesis space of relationships formed by pairs of entities. 
However, their densely-generated proposals often lead to high complexity in computation and a significant level of noise in context modeling (c.f., Sec.~\ref{sec:intro}).
Many heuristic strategies have been proposed to address these issues, such as improving proposal generation~\cite{yang_graph_2018} or efficient context modeling of graph structures using tree ~\cite{tang_learning_2018} or dynamic sparse graph~\cite{li_factorizable_2018, tang_learning_2018, qi_attentive_2018, yang2019auto, wang_exploring_2019, li2021bipartite}. 
Nonetheless, those strategies often require sophisticated model tuning and is difficult to achieve end-to-end optimization, which tends to limit the representation power of the proposed models.

\noindent\textbf{One-stage Scene Graph Generation}~
Inspired by the one-stage object detection methods~\cite{tian2019fcos,carion2020end,sun2021sparse}
, one-stage SGG approaches typically employ a fully-convolutional network~\cite{liu2021fully, teng2021structured} or a CNN-Transformer~\cite{dong2021visual, yang2022panoptic, cong2023reltr, khandelwal2022iterative} architecture to detect the relationship triplet directly from image features, treating each instance as a whole. 
Specifically, Liu \etal~\cite{liu2021fully} adopt the paradigm of fully-convolutional one-stage object detection FCOS~\cite{tian2019fcos}, which densely predicts the relationship triplets from image.
Tang \etal~\cite{teng2021structured} introduce a sparse set of entity-pair relation proposals, which can be viewed as a generalization of Sparse-RCNN\cite{sun2021sparse} for detecting visual relationships.
Along the line of CNN-Transformer architecture, Yang \etal~\cite{yang2022panoptic} extract relationships using the transformer model, which aggregates relation representation between relation queries and image features, while Cong \etal~\cite{cong2023reltr} proposes RelTR with two decoupled transformer decoder modules to extract the subject and object, respectively.
The ISG proposed by Khandelwal \etal~\cite{khandelwal2022iterative} employs three transformer decoders to sequentially process the subject, object, and predicate of visual relations from images.
These one-stage frameworks are more computationally efficient than their two-stage counterparts due to their use of sparse relationship proposals.
However, due to the lack of explicit entity-predicate compositional modeling, they often suffer from unreliable object detection, resulting in unsatisfactory performance for the complex visual relationships of real-world scenarios.
Moreover, the majority of one-stage methods ignore the scene graph consistency as they predict each relationship independently rather than generating a valid graph structure with consistent node-edge constraint.

\noindent\textbf{One-stage Human-Object Interaction}~
Our work is also related to the task of Human-Object Interaction (HOI) detection~\cite{li2021transferable}, which aims to predict a specific type of relationship triplets defined as <human, interaction, object>. 
Similar to the SGG task, 
there has been a recent trend towards investigating one-stage frameworks for Human-Object Interaction, typically utilizing an encoder-decoder architecture based on a fully-convolutional~\cite{liao2020ppdm, kim2020uniondet, wang2020learning} or Transformer network~\cite{zou2021end, chen2021reformulating, tamura2021qpic, kim2021hotr, zhang2021mining, zhang2022efficient, chen2021qahoi,park2022consistency, liao2022gen}.
While early works attempt to predict the interaction in a holistic manner, recent approaches take a divide-and-conquer strategy for detecting the human and interacted object. Most of them concentrate on composing triplets after decoding each component of the relationship. Particularly, Chen \etal~\cite{chen2021reformulating} presents two similar encoder-decoder architectures, which introduce a dual decoder to simultaneously extract the human, object and interaction, and then group the components into final triplets by a rule-based mechanism in inference time. Concurrently, Kim \etal~\cite{kim2021hotr} proposes a grouping strategy that can be trained end-to-end, atop the dual decoder architecture.
Another strategy is to represent the compositional structure before decoding the triplets:
Zhang \etal~\cite{zhang2022efficient, zhang2021mining} borrow the design of the two-stage pipelines by pairing detected entities and then constructing interaction queries.

{
Based on these designs, we propose a transformer-based bipartite graph generation framework to capture visual relationship composition. Our method uses an entity-aware predicate generator to explicitly model the entity-predicate relationship and a differentiable graph assembling mechanism to generate consistent graphs, unlike previous works. This design helps us represent the complex visual relationships in SGG tasks and improve computational efficiency.
}

%% file: sec/method.tex
\vspace{-0.3cm}
\section{Preliminary}
\vspace{-0.1cm}

In this section, we first introduce the problem setting of scene graph generation in Sec.~\ref{subsec:setting}, and then present an overview of our approach in Sec.~\ref{subsec:overview}.

\vspace{-0.25cm}
\subsection{Problem Setting}\label{subsec:setting}
\vspace{-0.1cm}

The task of scene graph generation aims to parse an input image into a scene graph $\mathcal{G}_{scene}=\{\mathcal{V}_e,\mathcal{E}_r\}$, where $\mathcal{V}_e$ is a set of nodes denoting noun entities and $\mathcal{E}_r$ is a set of edges that represent relationships or predicates between pairs of subject and object entities.
Specifically, each entity $v_i\in \mathcal{V}_e$ has a category label from a set of entity classes $\mathcal{C}_e$ and a bounding box depicting its location in the image, while each edge $e_{i\to j} \in \mathcal{E}_r $ between a pair of nodes $v_i$ and $v_j$ is associated with a predicate label from a set of predicate classes $\mathcal{C}_p$ in this task. 

In this work, we consider an equivalent form of the scene graph $\mathcal{G}_{scene}$ by explicitly representing the noun entities and the predicates as nodes while its edges indicate associations in relationship triplets~\cite{li2021bipartite}. Specifically, we define a bipartite graph consisting of two groups of nodes $\mathcal{V}_e,\mathcal{V}_p$, which correspond to entities and predicates, respectively. 
These two groups of nodes are connected by two sets of directed edges $\mathcal{E}_{e\rightarrow p}, \mathcal{E}_{p\rightarrow e}$ representing the order in the relationship triplets, i.e., from the subjects to predicates and the predicates to objects.
Hence the bipartite graph has a form as $\mathcal{G}_b=\{\mathcal{V}_e,\mathcal{V}_p, \mathcal{E}_{e\rightarrow p}, \mathcal{E}_{p\rightarrow e}\}$.
Given the bipartite graph representation, we then formulate the scene graph generation process as a bipartite graph construction task.

\vspace{-0.2cm}
\subsection{Method Overview}\label{subsec:overview}
\vspace{-0.1cm}

Our goal is to build an efficient bipartite graph generation process with unbiased relationship predictions. To this end, we consider a model family defined by a differentiable function $\mathcal{F}_{sgg}$ that takes an image $\mathbf{I}$ as the input and outputs the bipartite graph $\mathcal{G}_{b}$. This allows us to train the graph generation in a simple end-to-end manner and hence avoid the complexity in typical two-stage methods. More importantly, we
propose to explicitly model the bipartite graph construction process, which enables us to leverage the compositional property of relationships for more expressive representation learning.

\begin{figure*}[h]
	\centering
	\includegraphics[width=0.90\textwidth]{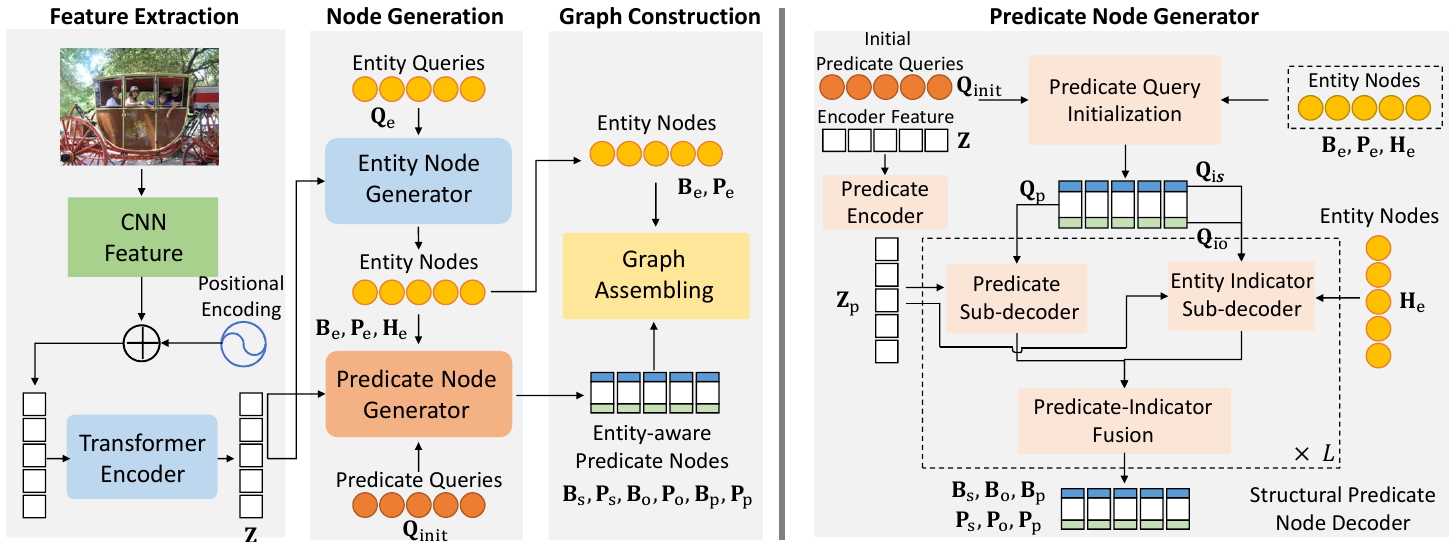}
	\vspace{-0.1cm}
	\caption{
		\textbf{An illustration of overall pipeline of our SGTR+ model.} \textbf{Left)} We use a CNN backbone together with a transformer encoder for image feature extraction. 
		The entity and predicate node generators are introduced to produce the entity node and entity-aware predicate node. 
		A graph assembling mechanism is developed to construct the final bipartite scene graph. 
		\textbf{Right)} The predicate node generator consists of three parts: \textit{a)} predicate query initialization, \textit{b)} a predicate encoder, and \textit{c)} a structural predicate node decoder, which is designed to generate entity-aware predicate nodes. 
	}
	\vspace{-0.55cm}
	\label{fig:main}
\end{figure*}

Specifically, our bipartite graph construction consists of two consecutive processes: {a) node~(entity and predicate) generation}, and {b) directed edge connection}.
In the {node generation} step, we extract the entity and predicate nodes from the image with an \textit{entity node generator} and a \textit{predicate node generator}, respectively.
For the {directed edge connection}, we design a \textit{graph assembling module} to predict the associations between the entity and predicate proposals.
An overview of our method is illustrated in left part of Fig.~\ref{fig:main}.
We have presented a base version of our method, i.e., SGTR, in our conference paper~\cite{li2022sgtr}. To further improve the learning of relationship triplets, we develop an enhanced model based on the SGTR framework, dubbed SGTR+, with a richer design in predicate node generator and more effective graph assembling. 
Below we will introduce details of our modules, including the original design of our SGTR framework and the improvements of SGTR+, followed by the model learning and inference.

\vspace{-0.45cm}
\section{Our Approach}\label{subsec:sgtr}
\vspace{-0.1cm}
In this section, we first introduce four main modules of our framework $\mathcal{F}_{sgg}$ using the base model SGTR as an example, which includes: 
(1) a \textit{backbone network} for generating feature representation of the scene (Sec.~\ref{subsubsec:backbone}); 
(2) a transformer-based \textit{entity node generator} for predicting entity proposals (Sec.~\ref{subsubsec:backbone});
(3) a structural \textit{predicate node generator} for producing predicate proposals (Sec.~\ref{subsec:triplet});
(4) a \textit{bipartite graph assembling} module for constructing final bipartite graph via connecting entity nodes with their corresponding predicate nodes (Sec.~\ref{subsec:Assembling}). 
Based on this framework, we introduce an enhanced and more general model design, \textit{SGTR+} in Sec.~\ref{subsec:sgtr+}, which has an efficient context-aware predicate representation and a principled graph assembling process.
Finally, the model learning and inference are detailed in Sec.~\ref{subsec:learning}.

\vspace{-0.36cm}
\subsection{Backbone Network and Entity Node Generator}\label{subsubsec:backbone}

We first use a CNN (e.g., ResNet) as the backbone network for extracting convolutional features from the image.
Motivated by the Transformer-based detector, DETR~\cite{carion2020end}, we then use a multi-layer Transformer encoder to encode convolutional features into a set of feature tokens. 
The resulting CNN-Transformer features are denoted as $\mathbf{Z} \in \mathbb{R}^{wh \times d}$, where $w,h,d$ are the width, height, and channel of the feature map, respectively. Those features are the inputs to subsequent entity and predicate node generator. 

For the entity node generator, we adopt the decoder of DETR to produce $N_e$ entity nodes from a set of learnable entity queries. 
Formally, we define the entity decoder as a mapping function $\mathcal{F}_{e}$, which takes initial entity query $\mathbf{Q}_{e}\in \mathbb{R}^{N_e\times d}$ and the feature tokens $\mathbf{Z}$ as inputs, and outputs the entity locations $\mathbf{B}_e\in{\mathbb{R}^{N_e\times 4}}$ and class scores $\mathbf{P}_e\in{\mathbb{R}^{N_e\times(\mathcal{C}_e+1)}}$, along with their associated feature representations $\mathbf{H}_e\in{\mathbb{R}^{N_e\times d}}$ as follows,
\vspace{-0.23cm}
\begin{align}
    \mathbf{B}_e,\mathbf{P}_e,\mathbf{H}_e = \mathcal{F}_{e}(\mathbf{Z}, \mathbf{Q}_{e}).
\end{align}
where $\mathbf{B}_e=\{\mathbf{b}_1,\cdots,\mathbf{b}_{N_e}\},\mathbf{b} = (x_c, y_c, w_b, h_b)$, $x_c, y_c$ are the normalized center coordinates, the $w_b,h_b$ are the width and height of each entity box.

\vspace{-0.2cm}
\subsection{Predicate Node Generator}\label{subsec:triplet}

Given the image feature $\mathbf{Z}$ and entity nodes, we now introduce our predicate node generator. Instead of extracting a set of predicate proposals separately, we intend to develop an entity-aware predicate representation that incorporates relevant entity information into each predicate node. 
Such a design allows us to encode potential associations between each predicate and its subject/object entities, which facilitates the graph assembling and efficient generation of the visual relation triplets. 

Concretely, our predicate node generator is composed of three components:
(1) a \textbf{predicate encoder} for relation-specific feature extraction (in Sec.~\ref{subsubsec:triplet_encoder}), 
(2) a \textbf{predicate query initialization} module for initializing a set of entity-aware predicate queries (in Sec.~\ref{subsubsec:query_generate}), and (3) a \textbf{structural predicate node decoder} for generating a set of entity-aware predicate nodes (in Sec.~\ref{subsubsec:triplet_decoder}). 
An overview of the predicate node generator is shown in Fig.~\ref{fig:main}-Right.  
Below, we will present the detailed design of the predicate node generator, starting with the SGTR in each component.

\vspace{-0.15cm}
\subsubsection{Predicate Encoder}\label{subsubsec:triplet_encoder}

In order to capture predicate-specific representation, we introduce a lightweight,Transformer predicate encoder to further refine the CNN-Transformer features $\mathbf{Z}$. 
It shares a similar structure as the the backbone Transformer encoder with fewer layers. 
The resulting predicate-specific feature is denoted as $\mathbf{Z}^p\in\mathbb{R}^{w\times h\times d}$.

\vspace{-0.1cm}
\subsubsection{Predicate Query Initialization}\label{subsubsec:query_generate}
We adopt a similar decoding strategy as the entity node generator, which generates predicate nodes from a set of learnable predicate queries. Instead of using a holistic vector-based query design, we propose a structural query representation that better fits into our graph construction process and enables us to more effectively exploit the compositional property of the visual relationships. 

Each predicate query comprises a \textit{predicate sub-query} and two \textit{entity indicator sub-queries} to indicate its association with subject and object entity proposals. This design enables the development of a structural predicate decoder (as detailed in Sec.~\ref{subsubsec:triplet_decoder}). 
The decoder predicts the predicate class/location and belief scores for its subject and object entities. These belief estimates significantly enhance the robustness of the subsequent graph assembly process, a critical step in learning a compositional representation for visual relationships.

Formally, we introduce a structural query representation with three components, denoted as $\mathbf{Q}_{p}^c=\{ \mathbf{Q}_{is}; \mathbf{Q}_{io};\mathbf{Q}_{p} \}\in \mathbb{R}^{N_r\times 3d}$, 
where $\mathbf{Q}_{is},\mathbf{Q}_{io} \in \mathbb{R}^{N_r\times d}$ represent the subject/object \textit{entity indicator sub-queries}, respectively\footnote{The subscripts 's', 'o' stand for the subject and object entity, respectively. }, and $\mathbf{Q}_{p} \in \mathbb{R}^{N_r\times d}$ is the \textit{predicate sub-query}. Given this structural design, we compute the predicate query $\mathbf{Q}_{p}^c$ in an entity-aware and scene-adaptive manner so that its components encode the corresponding information.

Because predicates are closely related to entities, we dynamically initialize the predicate queries based on the entities node in order to better encode the semantics of the predicate and the association between entities. This  is more effective than using shared global queries as in the original DETR. By using this dynamic initialization, we are able to more accurately capture the semantics of each predicate and its relationship between entities.

More specifically, we start from a set of initial predicate queries ${\mathbf{Q}}_{init} \in \mathbb{R}^{N_r\times d}$ and use the entity representation $\mathbf{B}_e, \mathbf{P}_e, \mathbf{H}_e$ to derive the query components $\mathbf{Q}_{is},\mathbf{Q}_{io},\mathbf{Q}_{p}$. 
To this end, we first compute an entity representation as in~\cite{yao2021efficient}, which defines a set of key and value vectors $\mathbf{K}_{init}, \mathbf{V}_{init}\in\mathbb{R}^{N_e\times d}$ as follows:
\vspace{-0.2cm}
\begin{align}
\mathbf{K}_{init}=\mathbf{V}_{init}=(\mathbf{H}_e+\mathbf{G}_e), \mathbf{G}_e=\text{ReLU}(\mathbf{B}_e\mathbf{W}_g).
\end{align}
where $\mathbf{G}_e\in\mathbb{R}^{N_e\times d}$ is a learnable geometric embedding of entity proposals, $\mathbf{W}_g\in\mathbb{R}^{4\times d}$ is a transformation from bounding box locations to the embedding space.
Given the augmented entity representations, we then compute the predicate queries $\mathbf{Q}_{p}^c$ using a multi-head cross-attention operation on the initial predicate queries ${\mathbf{Q}}_{init}$ and $\mathbf{K}_{init}$. 
\begin{align}
{\mathbf{Q}}^c_{p}&=\mathcal{A}({\mathbf{Q}}_{init},\mathbf{K}_{init},\mathbf{V}_{init}) \mathbf{W}_e , \\ 
\mathcal{A}(q, k, v)&=\text{FFN}(\text{MHA}(q, k, v)) .
\vspace{-1em}
\end{align}
where $\mathcal{A}(q, k, v)$ denotes the multi-head attention operation, and $\mathbf{W}_e \in\mathbb{R}^{d\times 3d}=[\mathbf{W}_e^{is},\mathbf{W}_e^{io},\mathbf{W}_e^{p}]$ are the transformation matrices for the three sub-queries~$\mathbf{Q}_{is}, \mathbf{Q}_{io}, \mathbf{Q}_{p}$, respectively. Such a cross-attention update fuses the entity information into the three query components, which are subsequently used in three respective sub-decoders for generating the predicate node.


\vspace{-0.2cm}
\subsubsection{Structural Predicate Node Decoder}\label{subsubsec:triplet_decoder}
Given the predicate query $\mathbf{Q}_{p}^{c}$, we now develop a structural predicate node decoder that generates a set of predicate proposals and their subject/object entity indicators.  
Here each predicate proposal consists of its predicate class and the center locations of its subject and object entities, and each entity indicator is represented as an entity-class distribution and its bounding box. Our predicate node decoder leverages the compositional property of visual relationships and computes the predicate proposal triplets from the entity and predicate features. 

To achieve this, we define a structural node decoder consisting of three modules: a) a \textit{predicate sub-decoder}; b) an \textit{entity indicator sub-decoder}; c) a  \textit{predicate-indicator fusion} module.
The first two modules of sub-decoders take the corresponding components of the predicate query and produce an updated representation of the predicate and entity indicators from the predicate feature $\mathbf{Z}^p$ and entity feature $\mathbf{H}_e$, respectively. To improve the entity-predicate association, the predicate-indicator fusion module then integrates the predicate and entity indicator representations, which are finally used to predict the predicate proposals and their entity indicators.    

Specifically, we use a multi-layer transformer for the two sub-decoders in our base version. For notation clarity, below we focus on a single decode layer and omit its layer index $l$ within each sub-decoder. We will re-introduce the layer index in the fusion module. 

\vspace{1mm}
\noindent\textit{A. Predicate Sub-decoder}.
The predicate sub-decoder is designed to generate a predicate representation from the predicate feature $\mathbf{Z}^p$ and the predicate sub-query $\mathbf{Q}_{p}$. It utilizes the image context via the cross-attention to update the predicate at each layer:
\vspace{-0.16cm}
\begin{align}
\widetilde{\mathbf{Q}}_{p}=\mathcal{A}(q=\mathbf{Q}_{p},k=\mathbf{Z}^p,v=\mathbf{Z}^p).
\end{align}
where the query is $\mathbf{Q}_{p}$, key and value are $\mathbf{Z}^p$, and 
$\widetilde{\mathbf{Q}}_{p}$ is the updated predicate representation.

\vspace{1mm}
\noindent\textit{B. Entity Indicator Sub-Decoder.}
The entity indicator sub-decoder generates a representation for the entity indicators based on the corresponding sub-queries and entity features, which encodes the associations between predicates and entities. 
Specifically, we perform a cross-attention operation between the entity indicator sub-queries
$\mathbf{Q}_{is},\mathbf{Q}_{io}$ and the entity proposal features $\mathbf{H}_e$ from the entity node generator. This enables us to encode the entity information in the representation of entity indicators.   
We denote the updated representation of the entities indicator as $\widetilde{\mathbf{Q}}_{is}, \widetilde{\mathbf{Q}}_{io}$, which are generated as follows:
\vspace{-0.17cm}
\begin{align}
    \widetilde{\mathbf{Q}}_{is}=\mathcal{A}(\mathbf{Q}_{is},\mathbf{H}_e,\mathbf{H}_e),\quad \widetilde{\mathbf{Q}}_{io} =\mathcal{A}(\mathbf{Q}_{io},\mathbf{H}_e,\mathbf{H}_e).
\end{align}

\vspace{1mm}
\noindent\textit{C. Predicate-Indicator Fusion.}\label{para:intr_refine}
To build the connection between the representations of each predicate and its entity indicators, 
we introduce a predicate-indicator fusion module to integrate the entity indicators into their predicate feature. Here we consider the $l$-th decoder layer, in which we fuse the features of three components $\widetilde{\mathbf{Q}}_{p}^l, \widetilde{\mathbf{Q}}_{is}^l, \widetilde{\mathbf{Q}}_{io}^l$ and produce an updated version of predicate query for the next layer, denoted as  $\mathbf{Q}^{l+1}_{p}, \mathbf{Q}^{l+1}_{is}, \mathbf{Q}^{l+1}_{io}$. 
Formally, we use a fully connected layer for updating the predicate sub-query and inherit the entity indicator representations as follows:
\vspace{-0.17cm}
\begin{align}
    \mathbf{Q}^{l+1}_{p} &= \widetilde{\mathbf{Q}}_{p}^l + \left(\widetilde{\mathbf{Q}}_{is}^l + \widetilde{\mathbf{Q}}_{io}^l\right) \cdot \mathbf{W}_{i} \label{eq:updatg_pred}, \\
    \mathbf{Q}^{l+1}_{is}&=\widetilde{\mathbf{Q}}_{is}^l,\quad \mathbf{Q}^{l+1}_{io}=\widetilde{\mathbf{Q}}_{io}^l.
\end{align} 
where $\mathbf{W}_{i}, \mathbf{W}_{p} \in \mathbb{R}^{d \times d}$ are the linear layer weights.

At each decode layer $l$, we use the fused features to generate the class and location predictions for predicates, as well as those of their entity indicators as follows,
\vspace{-0.18cm}
\begin{align}
    \mathbf{P}_p^l&=\sigma (\mathbf{Q}_{p}^l\cdot\mathbf{W}_{cls}^{p})\in\mathbb{R}^{N_r\times \mathcal{C}_p},\\ 
    \mathbf{B}_p^l&=\sigma({\mathbf{Q}}_{p}^l\cdot\mathbf{W}_{reg}^{p})=\{(x_c^s, y_c^s, x_c^o, y_c^o)\}\in\mathbb{R}^{N_r\times 4}.
\end{align}
where $\sigma$ is the sigmoid function, $\mathbf{P}_p^l$ are the predicted predicate class distributions, and $\mathbf{B}_p^l=\{(x_c^s, y_c^s, x_c^o, y_c^o)\}$ are the box center coordinates of their subject and object entities. 
The entity indicators are also decoded into their locations $\mathbf{B}_s^l,\mathbf{B}_o^l\in\mathbb{R}^{N_r\times 4}$ and object category predictions $ \mathbf{P}_s^l,\mathbf{P}_o^l\in\mathbb{R}^{N_r\times (\mathcal{C}_e+1)}$ in a similar manner.

We stack $L$ layers of those three modules, and each decoder layer gradually improves the quality of predicate and entity association. The predictions of the $L$-th layer is used as the final output of the predicate node decoder, which is fed into the subsequent graph assembling process.

\vspace{-0.2cm}
\subsection{Bipartite Graph Assembling}\label{subsec:Assembling}

\begin{figure}
    \centering
    \includegraphics[width=\linewidth]{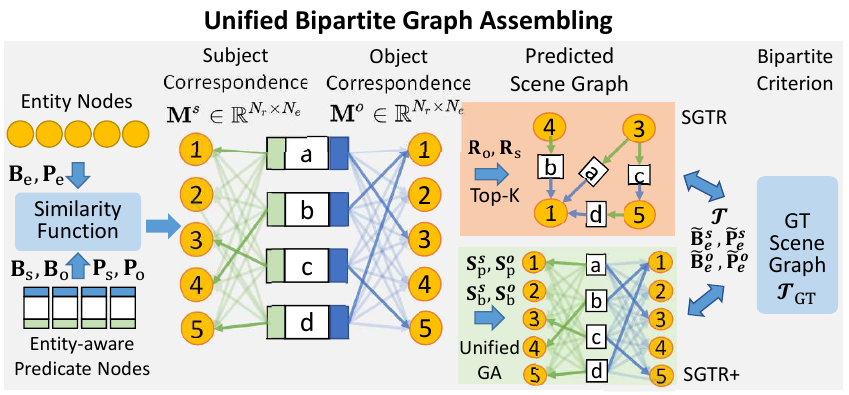}
    \vspace{-0.55cm}
    \caption{\textbf{The illustration of Unified Bipartite Graph Assembling.} 
    In SGTR, the partial differentiable top-k selection constructs relation triplets that match GT scene graphs (the upper orange block).
    The unified GA of SGTR+ (the lower green block) adopts weight sum to construct relation triplets, allowing GT to optimize all predicate entity associations.}
    \label{fig:modules} 
    \vspace{-0.56cm}
\end{figure}

Given the entity and predicate node proposals, we now introduce our bipartite graph assembling process. As mentioned before, we convert the original scene graph into a bipartite graph structure which consists of $N_e$ entity nodes and $N_r$ predicate nodes.
The main goal of the graph assembling is to link the entity-aware predicate nodes to the proper entity node. To this end, we consider two key steps in constructing the bipartite graph: (1) {calculating a correspondence score matrix} between $N_e$ entity nodes and $N_r$ predicate nodes; (2) {extracting relationship triplets} according to the node correspondence scores.
In each step, we first present a base version of our design as introduced in SGTR and then describe a unified strategy adopted in SGTR+. An overview of our bipartite graph assembling is shown in Figure.~\ref{fig:modules}. 

\vspace{-0.2cm}
\subsubsection{Node Correspondence Score}
Our first step is to estimate a correspondence score between each entity and predicate node pair. Taking the subject entity nodes as an example, we compute a correspondence score matrix  $\mathbf{M}^s\in\mathbb{R}^{N_r\times N_e}$, which measures the association between the $N_e$ subject entity nodes and $N_r$ predicate nodes. 
The association is based on a similarity score between the entity indicators of predicate nodes and the entity nodes. 

Specifically, we combine two similarity measures, including a categorical similarity, denoted as $\mathbf{S}^s_p \in \mathbb{R}^{N_r \times N_e}$, and a spatial location similarity, denoted as  $\mathbf{S}^s_b \in \mathbb{R}^{N_r \times N_e}$, to calculate the overall similarity score. Formally,  the correspondence matrices is obtained by  $\mathbf{M}^s = \mathbf{S}^s_p \circ \mathbf{S}^s_b$, where $\circ$ is Hadamard product. 
Our base model (i.e., SGTR) adopts two simple similarity functions to measure the matching quality from different dimensions, which are defined as follows,
\vspace{-0.25cm}
\begin{align}
 \mathbf{S}^s_b=d_{loc}(\mathbf{B}_{s},\mathbf{B}_e), \quad \mathbf{S}^s_p = d_{cls}(\mathbf{P}_{s},\mathbf{P}_e).
\end{align}
where $d_{loc}(\cdot)$ is the GIOU or $L_1$ distance between the bounding box predictions, and $d_{cls}(\cdot)$ are the cosine distance between the classification distributions\footnote{The details of the distance functions are presented in the Appendix.}.
The correspondence score matrix of object entity nodes,  $\mathbf{M}^o\in\mathbb{R}^{N_r\times N_e}$, is obtained in the same manner. 


\vspace{-0.25cm}
\subsubsection{Relationship Triplet Generation}

Given the node correspondence score matrices $\mathbf{M}^s$ and $\mathbf{M}^o$, we now describe the module for generating relationship triplets in scene graph. 
We denote the relationship triplet set as $\mathcal{T}=\{(\widetilde{\mathbf{b}}_{e}^s, \widetilde{\mathbf{b}}_{e}^o,\widetilde{\mathbf{p}}_{e}^s, \widetilde{\mathbf{p}}_{e}^o, \mathbf{p}_p,\mathbf{b}_p)\}$. Here $\widetilde{\mathbf{b}}_{e}^s, \widetilde{\mathbf{b}}_{e}^o\in\mathbb{R}^{1\times 4}$ and $\widetilde{\mathbf{p}}_{e}^s, \widetilde{\mathbf{p}}_{e}^o\in\mathbb{R}^{1\times (\mathcal{C}_e+1)}$ are the bounding boxes and class predictions of its subject and object entity respectively, $\mathbf{p}_p\in\mathbb{R}^{1\times (\mathcal{C}_p+1)}$ is the class prediction of each predicate, and $\mathbf{b}_p\in\mathbf{B}_p$ are the centers of the predicate's entities.

In our base model (i.e. SGTR), we build the relationship triplets by simply keeping the top-$K$ links according to the correspondence matrix:
\vspace{-0.24cm}
\begin{align}
\mathbf{R}^s &= \mathcal{F}_{top}(\mathbf{M}^s, K)\in\mathbb{R}^{N_r\times K},\\
\mathbf{R}^o &= \mathcal{F}_{top}(\mathbf{M}^o, K)\in\mathbb{R}^{N_r\times K}.
\end{align}
where $\mathcal{F}_{top}$ is the top-$K$ index selection operation, $\mathbf{R}^s$ and $\mathbf{R}^o$ are the index matrix of entities kept for each triplet from the two relationship roles of subject and object, respectively. 
Using the index matrix $\mathbf{R}^s$ and $\mathbf{R}^o$, we are able to generate the final relationship triplets as $\mathcal{T}$.

\begin{figure*}[t]
    \centering
    \includegraphics[width=0.92\linewidth]{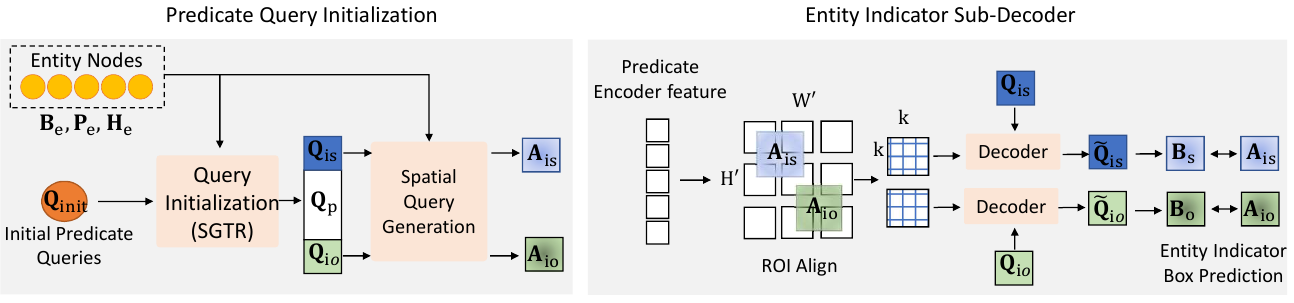}
    \vspace{-0.40cm}
    \caption{\textbf{The illustration of improved predicate node generator.} The left part illustrates the initialization of the spatial-aware predicate query. The right part illustrates the spatial-aware entity indicator sub-decoder.}
    \label{fig:sgtr+modules}
    \vspace{-0.6cm}
\end{figure*}

\vspace{-0.2cm}
\subsection{SGTR+}\label{subsec:sgtr+}
We now introduce an enhanced model design, SGTR+, in the SGTR framework, which improves two key modules: the predicate node generator and the graph assembling. In particular, we develop a \textit{spatial-aware predicate node generator} to generate an efficient predicate representation with image context and a differentiable \textit{unified graph assembling} to increase the quality of entity-predicate grouping.

\vspace{-0.15cm}
\subsubsection{Spatial-aware Predicate Node Generator}


\noindent\textbf{Spatial-aware Predicate Query}
For SGTR+, we add the spatial locations of entities to the entity indicator sub-query so that the existing entity nodes can be used to help with predicate modeling.
This design lets us use spatial clues from entity nodes, which improves the quality and efficiency of the next step decoding predicate nodes.

Specifically, we compute a \textit{spatial coordinate feature} for the entity indicator sub-queries, denoted as $\mathbf{A}_{is},\mathbf{A}_{io} \in \mathbb{R}^{N_r\times 4}$, which augment the original structural query $\mathbf{Q}_{p}^c$.
We demonstrate the process of query initialization in left part of Fig.~\ref{fig:sgtr+modules}.
In detail, we first compute an entity representation 
\vspace{-0.15cm}
\begin{align}
\mathbf{K}_{ent}= [\mathbf{B}_e,\mathbf{H}_e,\mathbf{P}_e]\mathbf{U}_e \in \mathbb{R}^{N_e \times d}.
\end{align}
based on the entity nodes, where $\mathbf{U}_e \in \mathbb{R}^{(4 + d + C_e) \times d}$ is a weight matrix. A spatial coordinate feature is derived from the bounding boxes of the entity proposals with an attention mechanism. Taking the subject entity indicator as an example, we have:
\vspace{-0.31cm}
\begin{align}
\mathbf{A}_{is} &=\mathcal{A}_{\tau}(\mathbf{Q}_{is}, \mathbf{K}_{ent}, \mathbf{B}_{e}), \\
&= \text{Softmax}\left( \frac{ d_{cos} (\mathbf{Q}_{is}, ~\mathbf{K}_{ent})}{ \tau } \right) \mathbf{B}_{e}, \\
d_{cos} (&\mathbf{Q}_{is}, \mathbf{K}_{ent}) = \frac{\mathbf{Q}_{is} \cdot\mathbf{K}_{ent}}{||\mathbf{Q}_{is}|| \cdot ||\mathbf{K}_{ent}||}.
\end{align}
where $d_{cos}(\cdot, \cdot)$ represents the cosine distance.
$\mathcal{A}_{\tau}$ is a simplified attention mechanism, employing a learnable parameter $\tau$ as temperature to modulate the softmax for aggregate coordinations from entity nodes.
It allows us to dynamically construct spatial representations of associated entities for predicate proposals originating from the entity nodes.

Our new predicate query $\mathbf{Q}^c_{p}$ comprises five components: ${ \mathbf{Q}_{is}, \mathbf{Q}_{io},\mathbf{Q}_{p}, \mathbf{A}_{is},\mathbf{A}_{io}}$, which is pivotal in the predicate decoding process, enabling the predicate decoder to capture the context of the predicate-entity relation effectively.

\noindent\textbf{Spatial-aware Predicate Node Decoder}
The explicit spatial cues $\mathbf{A}_{is}$ and $\mathbf{A}_{io}$ in the predicate proposal are used to guide the decoding of entity-aware information from the image features. 
These spatial coordinates enable the decoder to extract predicate-entity associations from the more expressive image-level predicate features $\mathbf{Z}^{p}$.
Moreover, this design significantly improves computational efficiency in contrast to the conventional use of entity features $\mathbf{H}_e$.

We now dive into the details of three essential components that constitute the predicate node decoder.

\vspace{1mm}
\noindent\textit{A. Predicate Sub-decoder}. SGTR+ shares the same predicate sub-decoder as the base version. 

\vspace{1mm}
\noindent\textit{B. Entity Indicator Sub-Decoder.} 
During the decoding of entity indicators in SGTR+, we utilize the spatial representations $\mathbf{A}_{is},\mathbf{A}_{io}$ to form anchor boxes, which guide the decoder in retrieving entity indicators from the image-level predicate feature $\mathbf{Z}^{p}$.
Specifically, we illustrate our design in the right part of Fig.~\ref{fig:sgtr+modules}, and will take the subject entity indicator as an example for the detailed description below. 
Concretely, we first use $\mathbf{A}_{is}$ to pool $N_r$ region features as candidate indicator features $\mathbf{Z}_e$ based on ROI-Align:
\vspace{-0.2cm}
\begin{align}
\mathbf{Z}_e &= \text{ROIAlign}(\mathbf{A}_{is}, \mathbf{Z}^{p} ) , \\ 
&=[\mathbf{z}_{e,1}, ...,\mathbf{z}_{e,N_r} ] \in \mathbb{R}^{N_r \times R \times d}.
\end{align}
where $R$ is the normalized size for each entity indicator anchor box.  
We then use the entity indicator sub-query $\mathbf{Q}_{is}$ to compute an indicator representation from $\mathbf{Z}_e$ in two steps. It starts with a self-attention operation on the sub-query, followed by a cross-attention between the updated sub-query $\mathbf{Q}_{is}^'$ and candidate feature $\mathbf{Z}_e$:
\vspace{-0.23cm}
\begin{align}
\mathbf{Q}_{is}^'  &= \mathcal{A}(\mathbf{Q}_{is} , \mathbf{Q}_{is} , \mathbf{Q}_{is} ), \\
\widetilde{\mathbf{Q}}_{is} &= \mathcal{A}(\mathbf{Q}_{is}^', \mathbf{Z}_{e} , \mathbf{Z}_{e}). 
\end{align}
The object entity indicator representation $\widetilde{\mathbf{Q}}_{io}$ is computed in a similar manner.
With the enhanced decoding and spatial coordinate features $\mathbf{A}_{is},\mathbf{A}_{io}$, the entity indicator decoding can be achieved with fewer decode layers compared with the base version (more detailed comparison in experiments).

\vspace{1mm}
\noindent\textit{C. Predicate-Indicator Fusion.} 
We adopt the same fusion procedure as the base version, while additionally updating the spatial coordinate features of entity indicator sub-queries based on decoding results: $\mathbf{A}_{is}^{l+1}= \mathbf{B}_s^l$, $\mathbf{A}_{io}^{l+1}=\mathbf{B}_o^l.$

\vspace{-0.15cm}
\subsubsection{Unified Graph Assembling}
Our new graph assembling module incorporates an adaptive \textit{learnable distance function} for the correspondence matrix and a fully differentiable \textit{unified triplets construction}, which improves stability in training and accuracy in inference.


\noindent \textbf{Learnable Distance Function}
The base SGTR uses a predefined distance function for the similarity between entity and predicate nodes, which is sensitive to varying entity indicator qualities.
To address this, we introduce similarity functions with learnable embeddings.
Specifically, we first compute a query and key embedding of the entity nodes, $\{\mathbf{P}_{e}, \mathbf{B}_{e}\}$, and the entity indicators,  $\{\mathbf{P}_{s}, \mathbf{B}_{s}\}$, as follows,
\vspace{-0.17cm}
\begin{align}
&\mathbf{K}_{e}^p = \mathbf{P}_e  \mathbf{W}_{pga}, \quad
\mathbf{Q}_{is}^p = \mathbf{P}_{s}  \mathbf{W}_{pga}, \;\mathbf{W}_{pga}\in\mathbb{R}^{C_e \times d},\\
&\mathbf{K}_{e}^b = \mathbf{B}_e  \mathbf{W}_{bga}, \quad
\mathbf{Q}_{is}^b = \mathbf{B}_{s}  \mathbf{W}_{bga} , \;\mathbf{W}_{bga}\in\mathbb{R}^{4 \times d}.
\end{align}
where $\mathbf{W}_{bga}, \mathbf{W}_{pga}$ are embedding weights. 

We then define the similarity functions based on the cosine distance between the key and value embeddings:
\vspace{-0.18cm} 
\begin{align}
    \mathbf{S}^s_p &= \text{softmax}\left( 
        \frac{ d_{cos} \left( \mathbf{Q}_{is}^p, \mathbf{K}_{e}^p\right) }
            {\tau_{ga}}\right),\\
    \mathbf{S}^s_b &= \text{softmax}\left(  
        \frac{  d_{cos} \left(\mathbf{Q}_{is}^b, \mathbf{K}_{e}^b\right)}
            {\tau_{ga}}\right).
\end{align}
where $\tau_{ga}$ represents a learnable temperature parameter controlling the sharpness of the similarity matrices. This learnable similarity function maps entity semantics into a hidden space, which can be jointly optimized with node generators, thereby accommodating low-quality entity nodes. 
The subsequent experimental section will provide an analysis of different distance metrics.

\noindent \textbf{Unified Triplets Construction}
The base SGTR employs a heuristic top-$K$ selection strategy to obtain triplets. 
However, during training, this often results in gradient instability and poor estimates of correspondence matrices. To tackle this, we introduce a \textit{unified triplets construction} for assembling relation triplets. 
Specifically, for model training, we design a soft prediction module to output entity box $\widetilde{\mathbf{B}}_{e}^s, \widetilde{\mathbf{B}}_{e}^s$ and logits $\widetilde{\mathbf{P}}_{e}^s, \widetilde{\mathbf{P}}_{e}^o$ of assembled triplets according to the similarity matrices. 
Concretely, we take a weighted sum from the entity node prediction $\mathbf{P}_e, \mathbf{B}_e$ based on the similarity matrices $\mathbf{S}^s_p, \mathbf{S}^s_b$ between entity nodes and predicate-entity indicators:
\vspace{-0.15cm}
\begin{align}
    \widetilde{\mathbf{P}}_{e}^s &= \mathcal{A}_{\tau} ( \mathbf{K}_{e}^p,\mathbf{Q}_{is}^p, \mathbf{P}_{e})= \mathbf{S}^s_p \cdot \mathbf{P}_e ,  \\ 
    \widetilde{\mathbf{B}}_{e}^s &= \mathcal{A}_{\tau} ( \mathbf{K}_{e}^b,\mathbf{Q}_{is}^b,\mathbf{B}_{e}) = \mathbf{S}^s_b \cdot \mathbf{B}_e.
\end{align}  
The outputs $\widetilde{\mathbf{P}}_{e}^s, \widetilde{\mathbf{B}}_{e}^s$ are directly used to calculate the loss with the ground-truth relations.

This fully differentiable design enables us to jointly optimize graph assembling module and node generators, which
leads to higher-quality relation predictions and provides us with better correspondence matrices for triplet construction.

\vspace{-0.25cm}
\subsection{Learning and Inference}\label{subsec:learning}
\vspace{-0.1cm}

\noindent\textbf{Learning}~
To optimize our SGTR model, we have developed a multi-task loss function that includes two components: $\mathcal{L}^{ent}$, which pertains to the entity node generator, and $\mathcal{L}^{prd}$, which pertains to the predicate node generator. The overall loss function is represented by: 
\vspace{-0.15cm}
\begin{align}
    \mathcal{L} = \mathcal{L}^{ent} + \mathcal{L}^{prd}.
\end{align} 
In our implementation, we use a DETR-like detector as entity node generator and the set-matching based loss function $\mathcal{L}^{ent}$, follows a similar form as described in previous work~\cite{carion2020end}. For more information about this loss function, please see the supplementary material. 

In the following sections, we will focus specifically on the loss function for the predicate node generator, $\mathcal{L}^{pre}$.
To calculate this loss, we first create a matching matrix between the predicted relationships (represented by $\mathcal{T}$) and the ground truth relationships (represented by $\mathcal{T}^{gt}$) using the Hungarian matching algorithm~\cite{kuhn1955hungarian}.
The cost of the set matching is defined as: 
\vspace{-0.15cm}
\begin{align}
    \mathcal{C} = \lambda_{p} \mathcal{C}_p + \lambda_{e} \mathcal{C}_e.
\end{align} 

The two components in the total cost correspond to the costs of predicate and subject/object entity, respectively
\footnote{ 
We utilize the location and classification predictions to calculate cost for each component. Detailed formulations are presented in supplementary.
}.
The matching index $\mathbf{I}^{tri}$ between triplet predictions and ground truths is produced by: $\mathbf{I}^{tri} = \text{argmin}_{\mathcal{T},\mathcal{T}^{gt}} \mathcal{C}$, which is used for following loss calculation of predicate node generator: $\mathcal{L}^{prd}=\mathcal{L}^{prd}_{i}+\mathcal{L}^{prd}_{p}+\mathcal{L}^{prd}_{a}$

The loss of predicate node generator $\mathcal{L}^{pre}$ has three terms $\mathcal{L}^{pre}_{i}, \mathcal{L}^{pre}_{p}, \mathcal{L}^{prd}_{a}$, which are used to supervise entity indicator sub-decoder, predicate sub-decoder in predicate node generator and unified graph assembling respectively.

For the loss of entity indicator sub-decoder $\mathcal{L}^{pre}_{i}$ and unified graph assembling $\mathcal{L}^{prd}_{a}$, we have,
\vspace{-0.14cm}
\begin{align} 
    \mathcal{L}^{prd}_i &= \mathcal{L}_{box}(\mathbf{B}_s, \mathbf{B}^s_{gt})+\mathcal{L}_{cls} (\mathbf{P}_s,\mathbf{P}^s_{gt}), \\ 
    \mathcal{L}^{prd}_a &= \mathcal{L}_{box}(\widetilde{\mathbf{B}}_{e}^s, \mathbf{B}^s_{gt}) + \mathcal{L}_{cls} (\widetilde{\mathbf{P}}_{e}^s, \mathbf{P}^s_{gt}). 
\end{align} 
where $\mathcal{L}^i_{box}$ and $\mathcal{L}^{i}_{cls}$ are the localization loss (L1 and GIOU loss) and cross-entropy loss for entities indicator $\mathbf{P}_s, \mathbf{B}_s,  \mathbf{P}_o, \mathbf{B}_o$ and entity of assembled triplet $\widetilde{\mathbf{B}}_{e}^s, \widetilde{\mathbf{B}}_{e}^o, \widetilde{\mathbf{P}}_{e}^s, \widetilde{\mathbf{P}}_{e}^o$.
\vspace{-0.14cm}
\begin{align}
    \mathcal{L}^{prd}_p = \mathcal{L}_{pos}^p(\mathbf{B}_p,\mathbf{B}^p_{gt}) + \mathcal{L}^p_{cls} (\mathbf{P}_p,\mathbf{P}^p_{gt}).
\end{align}
For the predicate sub-decoder, the $\mathcal{L}^p_{ent}$ is the L1 loss of the location of predicate's associated entities $\mathbf{B}_p$. 
The $\mathcal{L}^{p}_{cls}$ is the cross entropy of predicate category $\mathbf{P}_p$. 

In order to more effectively optimize our predicate node generator with a multi-layer decoder, we have implemented the auxiliary losses suggested in~\cite{carion2020end}. 
Specifically, we sum the predicate predictions from each layer, which are matched with the ground truth (GT), to calculate the predicate loss, denoted as $\mathcal{L}^{prd} = \sum_{l} \lambda_l \cdot \mathcal{L}^{prd}_l$. In this equation, $\lambda_l$ represents the loss weight for layer $l$. The anchor boxes of predicate queries are optimized by GT and matched with the first layer of the entity indicator sub-decoder.

\noindent\textbf{Inference}~
During inference, we use the top-$K$ links between entity and predicate nodes, as determined by the correspondence matrix, to construct the relationship triplets. This process is described in the base version of the graph assembly procedure.
After the assembly stage, we generate $K \cdot N_r$ visual relationship predictions. 

To improve the results even more, we remove from the scene graph any edges that aren't valid because they connect to themselves.
We use a post-processing step to remove self-connected triplets, which are triplets where the subject and object entities are the same.
We then rank the remaining predictions by the triplet score $\mathcal{S}^t = {(s^t_s \cdot s^t_o \cdot s^t_p)}$, where $s^t_s, s^t_o$, and $s^t_p$ represent the classification confidence of the subject entity $\widetilde{\mathbf{p}}_{e}^s$, object entity $\widetilde{\mathbf{p}}_{e}^o$, and predicate $\mathbf{p}_{p}$, respectively. From this ranked list of predictions, we select the top $N$ relationship triplets as our final outputs.

%% file: sec/exp.tex
\vspace{-0.3cm}
\section{Experiments}
\subsection{Experiments Configuration}
We evaluate our methods on Openimage V6 datasets~\cite{OpenImages}, Visual Genome~\cite{krishna2017visual} and GQA~\cite{hudson2018gqa}. 
We mainly adopt the data splits and evaluation metrics from the previous work~\cite{xu_scene_2017,zellers_neural_2017, li2021bipartite, knyazev2021generative}.
For the Openimage benchmark, the weighted evaluation metrics~($\text{wmAP}_{phr}$, $\text{wmAP}_{rel}$, score$_{wtd}$) are used for more class-balanced evaluation.
For the Visual Genome and GQA dataset, we adopt the evaluation metric recall@K~(R@K) and mean recall@K~(mR@K), harmonic recall@K~(hR@K) of SGDet, and also report the mR@100 on each long-tail category groups: \textit{head}, \textit{body} and \textit{tail} as same as~\cite{li2021bipartite}.

We use the ResNet-101, ResNet-50 and DETR~\cite{carion2020end} as backbone networks and entity detector, respectively.
To speedup training convergence, we first train entity detector on the target dataset, followed by joint training with predicate node generator.
For \textbf{SGTR}, the predicate node generator uses 3 layers of transformer encoder for predicate encoders and 6 layers of decoder for predicate and entity indicator sub-decoders.
For \textbf{SGTR+}, we use less predicate sub-decoder layer of 2 layers.
Both models' hidden dimensions $d$ is 256. 
Our predicate decoder uses $N_r$=200 queries. 
We set $K$=3 connection of correspondence matrices during inference for graph assembling module.
For more implementation details please refer to the supplementary.

\vspace{-0.35cm}
\subsection{Ablation Study}
\subsubsection{Model Components}\label{subsubsec:exp_model_comp_abla}
\begin{table}
    \begin{center}
        \resizebox{\linewidth}{!}{   
            \begin{tabular}{c|ccc|cccc}
                \toprule
               \#  & \textbf{EPN} & \textbf{SPD} & \textbf{GA}  & \textbf{mR@50} & \textbf{mR@100} & \textbf{R@50} & \textbf{R@100}  \\  \midrule
                1 & \cmark & \cmark & \cmark  & \textbf{13.9}  & \textbf{17.3} & \textbf{24.2}   & \textbf{28.2}       \\\midrule
                2 &        & \cmark & \cmark  & 12.0  & 15.9 & 22.9  & 26.3  \\
                3 & \cmark &        & \cmark  & 11.4  & 15.1 & 21.9  & 24.9   \\
                4 &        &        & \cmark  & 11.3  & 14.8 & 21.2  & 24.1   \\
                5 & \cmark & \cmark &         & 4.6   & 7.0  & 10.6   & 13.3     \\ 
                \toprule
            \end{tabular}
        }
    \end{center}
    \vspace{-0.35cm}
    \caption{\textbf{Ablation study on model components of SGTR.} EPN: Entity-aware Predicate Node; SPD: Structural Predicate Decoder, GA: Graph Assembling.} 
    \vspace{-0.25cm}
    \label{comp_abl_table}  
\end{table}

\begin{table}
    \begin{center}
        \resizebox{\linewidth}{!}{   
            \begin{tabular}{c|ccc|cc|c}
                \toprule
                \# & \textbf{SPQ} & \textbf{SPD} & \textbf{UGA} & \textbf{mR @ 50/100} & \textbf{R @ 50/100} & \textbf{Time/sec}  \\ \midrule 
                1 & \cmark   & \cmark   & \cmark   & \textbf{16.6}/\textbf{22.4}& \textbf{26.3}/\textbf{31.1}& 0.22     \\ \midrule
                2 &     &     &     & 13.9/17.3   & 24.2/28.2  & 0.35     \\
                3 & \cmark & \cmark &   & 13.6/19.9   & 25.5/28.7  & \textbf{0.19}     \\
                4 &     &     & \cmark & 15.3/19.2   & 26.2/31.0   & 0.36     \\ \midrule
                5 & &\cmark&\cmark & 15.3/19.8   & 26.2/31.0  & 0.24     \\
                6 & \cmark &  &\cmark & 14.2/17.7   & 25.2/29.3  & 0.37     \\
                7 & \cmark & Z &\cmark & 15.7/20.9   & 26.7/30.5  & 0.49     \\\midrule

                8 &     & \xmark &     & 9.8/13.1    & 23.7/27.8  & 0.20     \\  \bottomrule
                \end{tabular}
        }
    \end{center}
    \vspace{-0.3cm}
    \caption{\textbf{Ablation study on model components of SGTR+.} SPQ: Spatial-aware Predicate Query; SPD: Spatial-aware Predicate Decoder, UGA: Unified Graph Assembling.
    Z: Use image feature $\mathbf{Z}^p$ as input of predicate sub-decoder of SGTR.
    \xmark: Use 2 layer transformer for predicate sub-decoder.} 
    \vspace{-0.5cm}
    \label{tab:sgtr+_comp_abl_table}  
\end{table}

\noindent\textbf{SGTR}~
As shown in Tab.~\ref{comp_abl_table}, we ablate each module to demonstrate the effectiveness of our design on the validation set of Visual Genome.

\noindent$\bullet$ We find that using the holistic query for predicate rather than the proposed structural form decreases the performance by a margin of R@100 and mR@100 at \textbf{1.9} and \textbf{1.4} in row-2.

\noindent$\bullet$ Adopting the shared cross-attention between the image features and predicate/entity indicator instead of the structural predicate decoder leads to the sub-optimal performance as reported in row-3

\noindent$\bullet$ We further remove both entity indicators and directly decode the predicate node from the image feature.
The result is reported in row-4, which decreases the performance by a margin of \textbf{4.2} and \textbf{2.5} on R@100 and mR@100.

\noindent$\bullet$ In row-4, we directly use entity indicators as entity nodes for relationship prediction to study graph assembling. The model struggles to handle such complex multi-tasks within a single structure, but entity-prediction association modelling and graph assembling reduce optimization difficulty.

\noindent\textbf{SGTR+}~
To demonstrate the efficacy of our newly proposed SGTR+ design, we conduct experiments on the Visual Genome validation set.
We verify the spatial-aware predicate node generator and unified graph assembly with SGTR modules as the default structure of SGTR+.

\noindent$\bullet$ 
We shows effectiveness of SGTR+ by comparing newly proposed design, in Tab.~\ref{tab:sgtr+_comp_abl_table}, with SGTR module as base version.
We first module-level evaluate the technical improvement of SGTR+: \textit{spatial-aware predicate node generator} and \textit{unified graph assembling}. 
Compare row 3 with default SGTR in row 2, the \textit{spatial-aware node generator} achieves \textbf{2.6} improvement on mR@100, with less time complexity.

\noindent$\bullet$
The \textit{unified graph assembling design} also brings significant advantages for whole category space.
By comparing row-2 with row-4, newly proposed unified graph assembling module increases the performance by a margin of \textbf{1.9} and \textbf{2.8} on mR@100 and R@100, respectively.

\noindent$\bullet$
We examine the spatial-aware predicate node generator, introducing two innovations: spatial-aware queries and spatial-aware decoding from image features. We use learnable bounding boxes to initialize predicate queries from entity nodes for spatial-aware decoding, demonstrating the effectiveness of adding spatial coordinate features. This adjustment, implemented in row 5, results in a \textbf{2.6} decrease in mR@100 compared to row 1.

\noindent$\bullet$
To demonstrate the benefit of image feature for predicate decoding, we employ SGTR's standard transformer as an entity indicator sub-decoder to process the image-level predicate encoder feature $\mathbf{Z}^p$ in row-7. 
The $\mathbf{Z}^p$ demonstrated a \textbf{3.2} improvement on mR@100 compared to SGTR, which decodes from the entity node's hidden state (row-6).

\noindent$\bullet$
The time complexity of decoding from $\mathbf{Z}^p$  increased significantly due to the large size of the input data. The spatial-aware guided predicate decoder improves efficiency and node quality. A comparison between lines 1 and 7 shows a \textbf{1.5} margin on mR@100 and a \textbf{50\% reduction} in inference time (0.25 seconds), highlighting the advantages of our spatial-aware predicate decoder.

\noindent$\bullet$
We also compare SGTR+ to SGTR with the same decoder layer in row 8. 
Without efficient decoding modelling and unified graph assembling, this light-weight SGTR achieves much lower performance on the SGG task.

\begin{table}
    \centering
        \resizebox{0.41\textwidth}{!}{        
            \begin{tabular}{l|l|cccc}
                \toprule
                \# & \textbf{GA} & \textbf{mR@50} & \textbf{mR@100} & \textbf{R@50} & \textbf{R@100}  \\ \midrule
                1 & SP & 10.6    & 11.8   & 24.4   & 27.7            \\
                2 & FT  &  13.3   & 16.1  & 23.7  & 27.5  \\ \midrule
                3 & \textbf{PDG} & 13.9  & 17.3  & 24.2 & 28.2   \\ 
                4 & \textbf{UGA} & \textbf{15.3}  & \textbf{19.2}  & \textbf{26.2} & \textbf{31.0}   \\ \midrule
                5 & PDG-G  &  4.6   & 5.5  & 7.9  & 11.3  \\ 
                6 & UGA-G  &  0.3   & 3.4  & 3.5  & 5.7  \\ 
                \bottomrule
                \end{tabular}
        }
        \vspace{-0.20cm}
        \caption{\textbf{Ablation study on graph assembling based on SGTR}. \textbf{SP}: spatial distance predicate-entity matching function proposed by AS-Net\cite{chen2021reformulating}; \textbf{FT}: feature similarity-based matching function proposed by HOTR~\cite{kim2021hotr};
        \textbf{PDG}:Pre-defined graph assembling used in SGTR;
        \textbf{UGA}:Unified graph assembling used in SGTR+;
        \textbf{-G}: Gumbel softmax for graph assembling.} 
        \label{ga_abl_table} 
        \vspace{-0.6cm}
\end{table}

\vspace{-0.3cm}
\subsubsection{Graph Assembling Design}
\noindent\textbf{Comparison with Existing Works}~
We further investigate the effectiveness of graph assembling design with existing works on the validation set of Visual Genome in Tab.~ \ref{ga_abl_table}. 
For more comprehensive comparison, we adopt the differentiable human-object pair matching function proposed by recent HOI methods~\cite{chen2021reformulating, kim2021hotr}.

\noindent$\bullet$
AS-Net\cite{chen2021reformulating} categorises entities depending on the interaction branch's predicted distance between entity bounding box and centre, without semantic information.
In feature space, HOTR~\cite{kim2021hotr}  measures predicate-entity cosine similarity. This form calculates the distance between the entity indicator $\widetilde{\mathbf{Q}}_{is}$, $\widetilde{\mathbf{Q}}_{io}$ and entity nodes $\mathbf{H}_e$.
Our graph assembling mechanism, which incorporates semantic and spatial information, is better than location-only~\cite{chen2021reformulating} and feature-based similarity~\cite{kim2021hotr}. 

\noindent$\bullet$
Our new unified graph assembling design, which uses feature-based matching design and prior-based matching strategy, optimizes overall and per-class metrics. 
Training convergence speed and stability improve. We will discuss it in the following section of qualitative analysis.

\noindent$\bullet$
{
We also use Gumbel softmax with pre-defined distance function (row-5) and parameterized distance function (row-6).
The results show that Gumbel tricks cannot provide stable gradient for jointly optimizing the node generator and graph assembling module, even with differentiable select operation.
}

\noindent$\bullet$
{
We conducted an in-depth analysis of the properties of the parameter $\tau$ in the context of graph assembling. Our investigation encompassed its evolution during the training process and the implications of various initialization choices. 
For more detail of these aspects, please refer to the supplementary material.
}

\begin{table}
    \begin{center}
        \resizebox{0.50\textwidth}{!}{      
            \begin{tabular}{c|l|cc|ccc|c}
                \toprule
                 &\#L & \textbf{mR@50/100} & \textbf{R@50/100} & \textbf{Head} & \textbf{Body} & \textbf{Tail} & \textbf{Time} \\ \midrule
                 \multirow{5}{*}{\rotatebox{90}{ND}} & 1 & 13.6~/~18.0 & 25.8~/~29.3 & 28.5 & 22.4 & 10.5 & 0.19 \\
                & \textbf{2} & \textbf{16.6}~/~\textbf{22.4} & \textbf{26.3}~/~\textbf{31.1}  &  \textbf{29.4} &  \textbf{22.5} &  \textbf{19.1} & 0.22  \\
                & 3 & 16.4~/~21.0 & 26.9~/~30.6 & 29.8  & 23.4 &  16.0 & 0.26 \\
                & 6 & 14.7~/~18.7 & 26.0~/~30.0 & 29.2 & 21.7 & 12.6  & 0.31 \\
                & 6* & 15.9~/~21.5 & 26.6~/~30.9 & 29.1 & 22.3 & 17.3 & 0.31 \\ \midrule
                \multirow{4}{*}{\rotatebox{90}{NE}} & 0 & 13.4~/~17.2 & 26.4~/~30.2 & 28.2 & 21.4 & 11.2 & 0.16  \\
                & 1 & 15.6~/~21.0 & 26.1~/~30.8 & 29.0 & 22.3 & 18.3 & 0.18 \\
                & \textbf{3} & \textbf{16.6}~/~\textbf{22.4} & 26.3~/~\textbf{31.1} & \textbf{29.4} & 22.5 & \textbf{19.1} & 0.22  \\
                & 6 & 16.4~/~22.1 & \textbf{26.4}~/~30.9 & 29.1 & \textbf{22.6} & 18.9 & 0.30  \\
                \bottomrule
                \end{tabular}
            }
    \end{center}
    \vspace{-0.35cm}
    \caption{\textbf{Ablation study on the number of decoder layers of the predicate node generator of SGTR+.} ND / NE: number of decoder/encoder layers for predicate node generator.
    *: denote the earlier stopping.} 
    \label{tab:sgtr+model_scale_table} 
    \vspace{-0.55cm}
\end{table}

\vspace{-0.1cm}
\subsubsection{Model Size}

\noindent\textbf{Predicate Node Generator}
We incrementally varied the number of layers ($L$) in the predicate decoder and encoder from $L=1$ to $L=6$, as shown in Tab. ~\ref{tab:sgtr+model_scale_table}.
The results show that the performance is saturated when $L=2$.
It was observed that increasing the decoder layers led to overfitting within the same training iterations. Consequently, the early stopping results in performance on par with SGTR+ using a 2-layer decoder, as shown in row 5.
Compared to the optimal choice in $L=6$ in SGTR\cite{li2022sgtr}, the spatial-aware node generator improves efficiency and reduces model complexity, resulting in better performance.
We also validate the effectiveness of the predicate encoder.
As shown in the lower part of Tab.~\ref{tab:sgtr+model_scale_table}.
With a sufficient layer of predicate encoder, the further refined feature enhances predicate modeling on the whole category space.

\noindent\textbf{Entity Detector}
As we adopt different entity detectors compared to previous two-stage designs, we conduct experiments to analyze the influence of detectors on the SGTR. The detailed results are presented in the supplementary.

\begin{table}
    \begin{center}
        \resizebox{0.46\textwidth}{!}{        
            \begin{tabular}{l|l|cc|cc|c}
                \toprule 
            \multirow{2}{*}{\textbf{B}} & \multirow{2}{*}{\textbf{Models}} &\multirow{2}{*}{\textbf{mR@50}} &  \multirow{2}{*}{\textbf{R@50}} &\multicolumn{2}{c|}{\textbf{wmAP}} 
             & \multirow{2}{*}{\textbf{score}\scriptsize{wtd}}    \\ 
             \cmidrule{5-6}
             &  &  &   & \textbf{rel}& \textbf{phr} &    \\ 
                \midrule
                \multirow{3}{*}{\rotatebox{90}{X101-F}} 
                     & RelDN   & 37.20  & \textbf{75.40} & 33.21 & 31.31   & 41.97   \\
                     & GPS-Net & 38.93  & 74.74 & 32.77 & 33.87  & 41.60   \\
                     & BGNN    & 40.45  & 74.98 & 33.51  & 34.15   & 42.06  \\ \midrule
                \multirow{8}{*}{\rotatebox{90}{R101}}     
                & BGNN$^{*\dagger}$  & 39.41 & 74.93  & 31.15 & 31.37 &  40.00  \\
                & RelDN$^{\dagger}$ & 36.80 & 72.75  & 29.87 & 30.42 &  38.67  \\ \cmidrule{2-7}
                & HOTR$^{\dagger}$   & 40.09 & 52.66 & 19.38 & 21.51 &  26.88   \\
                & AS-Net$^{\dagger}$ & 35.16 & 55.28 & 25.93 & 27.49 &  32.42 \\
                & RelTR & 29.34 & 64.47 & 34.17 & 37.44 &  41.54 \\
                & \textbf{SGTR}  & 42.61 & 59.91 & 36.98 & 38.73 &  42.28  \\
                & \textbf{SGTR+}  & \textbf{42.70} & 72.16 & \textbf{39.54} & \textbf{41.53} &  \textbf{45.59}  \\  \bottomrule
                \end{tabular}

        }
    \end{center}
\vspace{-0.36cm}
\caption{\textbf{The Performance on Openimage V6.} 
        $\dagger$ denotes results reproduced with the authors' code. The performance of ResNeXt-101 FPN is borrow from \cite{li2021bipartite}. * means using resampling strategy.} 
        \vspace{-0.63cm}
        \label{oiv6_overall_table} 
\end{table}

\vspace{-0.3cm}
\subsection{Comparisons with State-of-the-Art Methods} \label{subsec:sota_comp}
\vspace{-0.1cm}

\begin{table*}[!ht]
    \begin{center}
        \resizebox{0.8\textwidth}{!}{        
                \begin{tabular}{l|l|l|l|ccc|ccc|c}
                \toprule
                \textbf{B} & \textbf{D} & \textbf{L}    & \textbf{Method}       & \textbf{mR@50/100} & \textbf{R@50/100} & \textbf{hR@50/100} & \textbf{Head}  & \textbf{Body}  & \textbf{Tail}  & \textbf{Time/Sec} \\ \midrule
                $\star$ & $\star$ &  & FCSGG  \cite{liu2021fully}  & 3.6~/~4.2 & 21.3~/~25.1 & 6.1~/~7.2 &-&-&- & 0.12 \\\midrule
                \multirow{10}{*}{\rotatebox{90}{X101-FPN}} & \multirow{9}{*}{\rotatebox{90}{Faster-RCNN}} 
                                          & & RelDN \cite{li2021bipartite}     & 6.0~/~7.3   & 31.4~/~35.9 & 10.1~/~12.1 & - & - & - & 0.65  \\
                                         & & & Motifs\cite{tang_unbiased_2020}     & 5.5~/~6.8 &  \textbf{32.1}~/~\textbf{36.9} &  9.4~/~11.5 & -  & - & - & 1.00   \\
                                         & & & VCTree\cite{tang_unbiased_2020}     & 6.6~/~7.7 &  31.8~/~36.1 & 10.9~/~12.7 & -  & - & - & 1.69   \\
                                         & & & GPS-Net\cite{lin_gps-net_2020, li2021bipartite}     & 6.7~/~8.6 &  31.1~/~35.9 & 11.0~/~13.8  & -  & - & - & 1.02   \\ \cmidrule{3-11} 
                                         & & R$_{\text{LVIS}}$ \cite{gupta_lvis:_2019}& GPS-Net\cite{li2021bipartite}$^\dagger$ &  7.4~/~9.5& 27.8~/~32.1 & 11.8~/~14.6 & 32.3 & 9.9 & 4.0 & $\ge$1.69   \\
                                         & & R$_{\text{BiLvl}}$\cite{li2021bipartite}& BGNN \cite{li2021bipartite}   & 10.7~/~12.6 & 31.0~/~35.8 & 15.9~/~18.6  & 34.0 & 12.9 &  6.0 & 1.32\\ 
                                         & & L$_{\text{TDE}}$ \cite{tang_unbiased_2020}  & VCTree\cite{tang_unbiased_2020}        &  9.3~/~11.1 &  19.4~/~23.2 & 12.6~/~15.0 & - & - & - & $\ge$1.69   \\
                                         & & L$_{\text{BPLSA}}$ \cite{guo2021general} & VCTree \cite{guo2021general}          & 13.5~/~15.7 & 21.7~/~25.5 &  16.6~/~19.4 & - & - & - &$\ge$1.69 \\
                                         \cmidrule{2-11} 
                                          & \multirow{2}{*}{$\square$}  &  & SSRCNN \cite{teng2021structured}  & 8.6~/~10.3 & \textbf{33.5}~/~\textbf{38.4} & 13.6~/~16.2 & 35.7 & 9.7 & 2.1 & 0.41 \\
                                           & & L$_{\text{LA}}$ \cite{menon2020long} & SSRCNN \cite{teng2021structured}  & \textbf{18.6}~/~\textbf{22.5} & 23.7~/~27.3 & \textbf{20.8}~/~\textbf{24.6}& 29.3 & 23.9 & 18.7 & 0.41 \\
                                          \midrule
                \multirow{12}{*}{\rotatebox{90}{R101}} & \multirow{18}{*}{\rotatebox{90}{DETR}}
                                          & & RelDN$^{\dagger}$ \cite{li2021bipartite}  & 4.5~/~5.2  & 24.7~/~27.3 & 7.7~/~9.3 & 26.9 & 5.8  & 0.9 & 0.65\\ 
                                          & && AS-Net$^{\dagger}$  \cite{chen2021reformulating} & 6.1~/~7.2   & 18.7~/~21.1 & 92~/~10.7 & 19.6 & 7.7 & 2.7 & 0.33   \\
                                          & && HOTR$^{\dagger}$ \cite{kim2021hotr}     & 9.4~/~12.0  &  23.5~/~27.7 & 13.4~/~16.7 & 26.1 & 16.2 & 3.4 & 0.25   \\
                                          &  & R$_{\text{BiLvl}}$ \cite{li2021bipartite} & BGNN &  6.3 / 7.7   &   25.4 / 29.5  &  13.2~/~15.8 & 28.5  & 7.2  & 1.5 & 0.83 \\ 
                                          \cmidrule{3-11}
                                          & && SGTR    & \underline{12.0}~/~\underline{15.2}    & 24.6~/~28.4  & 16.1~/~19.8 & 28.2 & 18.6 & 7.1 & 0.35      \\
                                          & && ISG\cite{khandelwal2022iterative} & 8.0~/~8.8 &  \textbf{29.7}~/~\textbf{32.1} & 12.6~/~13.8 & \textbf{31.7} & 9.0 & 1.4 & \underline{0.31} \\
                                          & && \textbf{SGTR+} & \textbf{13.1}~/~\textbf{17.0} &  \underline{26.4}~/~\underline{30.1}  & \textbf{17.5}~/~\textbf{21.7} & \underline{29.1}  & 21.7 & 8.8 & \textbf{0.22} \\   \cmidrule{3-11}
                                          & & R$_{\text{BiLvl}}$ \cite{li2021bipartite} & SGTR &  15.8~/~20.1 & 20.6~/~25.0 & 17.9~/~22.3 & 21.7 & 21.6 & \underline{17.1}& 0.35   \\  
                                          & & R$_{\text{BiLvl}}$ \cite{li2021bipartite} &  ISG$^{\dagger}$ & 11.3~/~14.6 &  \textbf{23.6}~/~\textbf{28.2} & 15.3~/~19.2 & 28.1 & 16.4 & 5.7 & 0.31   \\
                                          & & R$_{\text{BiLvl}}$+L$_{\text{RW}}$\cite{khandelwal2022iterative} &  ISG & \underline{17.1}~/~19.2 &  22.9~/~25.7 & \textbf{19.6}~/~22.0 & 24.4 & 20.2 & 16.4 & \underline{0.31}   \\ 
                                          & & R$_{\text{BiLvl}}$ \cite{li2021bipartite} &  \textbf{SGTR+} & 16.6~/~\underline{21.0} &  \underline{23.5}~/~\underline{27.3} & \underline{19.4}~/~\textbf{23.7} & 26.0 & \underline{23.9} & 16.1 & \textbf{0.22}   \\
                                          & & R$_{\text{BiLvl}}$+L$_{\text{LA}}$  &  \textbf{SGTR+} & \textbf{17.7}~/~\textbf{22.2} &  20.4~/~25.3 &  19.0~/~\underline{23.6} & 23.7 & \textbf{25.5} & \textbf{19.0} & \textbf{0.22}  \\
                                          \cmidrule{1-1}  \cmidrule{3-11}
                    \multirow{5}{*}{\rotatebox{90}{R50}}     
                                        & && RelTR\cite{cong2023reltr} & 8.4~/~9.8 &  \underline{25.2}~/~\underline{27.4} & 12.6~/~14.4  & \underline{27.9} & 10.0 & 3.2 &  \textbf{0.16}  \\
                                        & && SGTR & 11.5~/~15.7 &  22.9~/~25.8 & 15.3~/~19.5 & 25.5 & 20.0 & 8.5 &  0.26  \\
                                        & &&  \textbf{SGTR+} & 12.6~/~17.0 &  \textbf{25.8}~/~\textbf{29.6} & 16.9~/~\underline{21.6} & \textbf{28.7} & 21.2 & 9.2 &  \underline{0.18} \\ 
                                        & & R$_{\text{BiLvl}}$ \cite{li2021bipartite} & SGTR & \underline{15.1}~/~\underline{18.9} &  20.8~/~23.6 & \underline{17.5}~/~21.0 & 22.9 & \underline{21.5} & \underline{15.1} & 0.26   \\
                                        & & R$_{\text{BiLvl}}$ \cite{li2021bipartite} &  \textbf{SGTR+} & \textbf{16.4}~/~\textbf{20.3} &  22.9~/~{26.6} & \textbf{19.1}~/~\textbf{23.0} & 25.4 & \textbf{23.4} & \textbf{15.6} &  \underline{0.18} \\
                                          \bottomrule
                \end{tabular}
        }
    \end{center}
    \vspace{-0.28cm}
    \caption{
        \textbf{The SGDet performance on test set of Visual Genome dataset.} 
        \textbf{B} denotes the type of backbone, \textbf{D} denotes the type of detector, the \textbf{L} denotes the longtail strategies.
        Specifically, the 'R' is the training time data resampling strategy, and 'L' is the logit adjustment during the inference time.
        $\dagger$ denotes results reproduced with the authors' code. 
        $*$ denotes the bi-level resampling \cite{li2021bipartite} is applied for this model. 
        $\star$ denotes the special backbone  HRNetW48-5S-FPN$_{\times \text{2-f}}$ 
        and entities detector, CenterNet\cite{zhou2019objects}.
        $\square$ denotes the one-stage object detector: sparse-RCNN \cite{sun2021sparse}.
        The best result is highlighted by \textbf{bold}, second one is marked by \underline{underline}.
    } 
    \vspace{-0.58cm}
\label{vg_overall_table}
\end{table*}

We conduct experiments on Openimage-V6, Visual Genome and GQA dataset to demonstrate the effectiveness of our design. 
We compare our method with several state-of-the-art two-stage(\textit{e.g.}, VCTree-PCPL, VCTree-DLFE, BGNN~\cite{li2021bipartite}, VCTree-TDE and one-stage SGG methods(\textit{e.g.} FCSGG~\cite{liu2021fully}, ISG~\cite{khandelwal2022iterative}, RelTR~\cite{cong2023reltr}).
Furthermore, we compare with several strong one-stage HOI methods with similar entity-predicate pairing mechanisms (AS-Net~\cite{chen2021reformulating}, HOTR~\cite{kim2021hotr}) using their released code for a more comprehensive comparison.
Since our backbone is different from previous two-stage SGG (ResNeXt-101 with FPN), we reproduced the SOTA methods BGNN and its baseline RelDN with the same ResNet-101 backbone for more fair comparisons.
Besides, towards to more lightweight SGG model, we also evaluate our models upon on ResNet-50 for less time consumption.

\noindent\textbf{OpenImage V6}~
In Tab.~\ref{oiv6_overall_table}, we present our results on the OpenImage V6 dataset using ResNet-101. Our methods, SGTR and SGTR+, surpass the two-stage SOTA method BGNN by margins of \textbf{2.28} and \textbf{5.59}, respectively.
Additionally, SGTR shows notable improvements in weighted mAP metrics for relationship detection ($\text{wmAP}_{rel}$) and phrase detection ($\text{wmAP}_{phr}$), with gains of \textbf{5.83} and \textbf{7.36}, respectively.
Moreover, SGTR+ demonstrates a substantial enhancement in recall R@50 by \textbf{21.25}, highlighting the benefits of entity modeling for relationship modeling.

\begin{figure}[h!]
    \centering
    \vspace{-0.43cm}
    \includegraphics[width=0.65\linewidth]{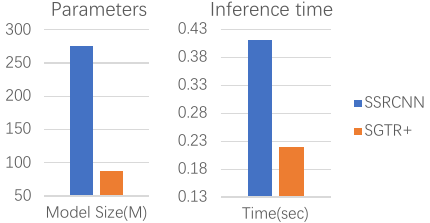}
    \vspace{-0.30cm}
    \caption{\textbf{The complexity comparison with SSRCNN.}}
    \label{fig:ssrcnn-cmp}
    \vspace{-0.3cm}
\end{figure}

\begin{figure*}[h]
    \centering
    \includegraphics[width=0.89\linewidth]{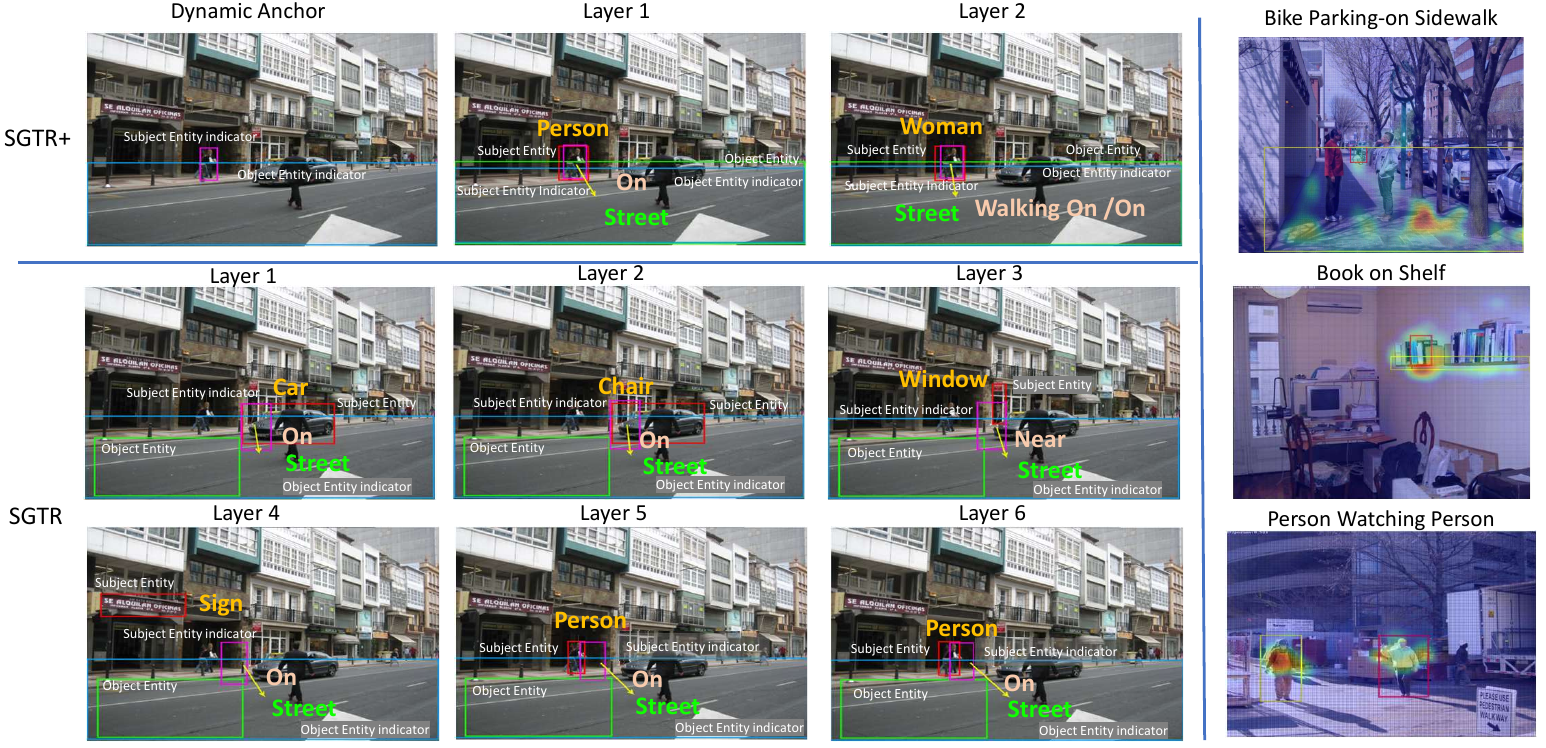}
    \vspace{-0.25cm}
    \caption{\textbf{Visualization of layer-wise predictions of SGTR and SGTR+ and attention of the predicate decoder of SGTR+.}
     The \textcolor{red}{\textbf{red}} and \textcolor{magenta}{\textbf{magenta}} boxes represent subject entity nodes and entity indicators, while the \textcolor{green}{\textbf{green}} and \textcolor{blue}{\textbf{blue}} boxes represent object entity nodes and entity indicators, respectively. The \textcolor{yellow}{\textbf{yellow}} arrow denotes the predicate interaction vector indicating subject-object associations. Best viewed in color.
    }
    \label{fig:lyr_pred_cmp}
    \vspace{-0.49cm}
\end{figure*}

\noindent\textbf{Visual Genome}~
As shown in Tab. \ref{vg_overall_table}, with the same ResNet-101 backbone,  we compare our method with the two-stage method BGNN~\cite{li2021bipartite}, and the one-stage methods HOTR~\cite{kim2021hotr}, AS-Net~\cite{chen2021reformulating}.
It shows that our method outperforms HOTR with a significant margin of \textbf{4.9} and \textbf{3.2} on mRecall@100.
We also use the ResNet-50 backbone as a more lightweight version of SGTR+. The results shows that our design decreases about \textbf{50\%} inference time with performance improvements.

\noindent$\bullet$ 
The sparse proposal set of SGTR+ achieves a more balanced foreground/background proposal distribution compared to traditional two-stage designs, reducing negative samples due to exhausted entity pairing. 
As a result, our method achieves competitive mean recall with the same backbone and learning strategy as shown in \cite{li2021bipartite}. 
We also introduce a novel long-tail recognition strategy \cite{menon2020long}, and inference-time logit adjustment further enhances mR@100 with minimal recall drop.

\noindent$\bullet$
Compared to the strong baseline SGTR, SGTR+ improves m@100 and R@100 by \textbf{1.8} and \textbf{1.7}, respectively, without using the longtail technique. The added resampling strategy further enhances performance, with improvements of \textbf{0.9} and \textbf{2.3} on m@100 and R@100. These findings highlight SGTR+'s superior trade-off between overall and mean recall in longtail scenarios. Notably, SGTR+ with a lightweight ResNet-50 backbone rivals the performance of ResNet-101, showcasing the effectiveness of entity-aware enhancement.

\noindent$\bullet$ 
However, our model underperforms in the head category compared to two-stage methods with similar backbones and one-stage methods with stronger detectors and backbones. This is partly due to the visual genome dataset containing many small entities, which our method struggles to recognize. 
We also evaluated the state-of-the-art two-stage method BGNN using DETR as a detector, which resulted in lower recall. For a more detailed limitation analysis, please refer to the supplementary material.

\begin{table}[h!]
    \centering
    \vspace{-0.15cm}
        \resizebox{0.40\textwidth}{!}{        
            \begin{tabular}{l|cccc}
                \toprule
                 zR &  {BGNN}$^{\dagger} $ & {VCTree-TDE} &  {\textbf{SGTR}} &  {\textbf{SGTR+}} \\ \midrule
                @50  & 0.4 & 2.6  & 2.4 & \textbf{3.8}  \\
                @100 & 0.9 & 3.2 &  \textbf{5.8}  & \textbf{5.8}\\ \bottomrule
                \end{tabular}
        }
        \vspace{-0.18cm}
        \caption{The zero-shot performance on VG.} 
        \label{tab:zero_shot} 
        \vspace{-0.50cm}
 \end{table}
 
\noindent$\bullet$
For zero-shot performance on VG, we compare SGTR with strong baseline: VCTree-TDE in Tab.\ref{tab:zero_shot}, achieving a gain of \textbf{2.6} on zR@100.
Our SGTR+ has further \textbf{1.4} improvement on zR@50.
The result shows the generalization capability of SGTR family to unseen relationships.

\noindent$\bullet$
We compared SGTR's efficiency to previous methods on an NVIDIA GeForce Titan XP GPU, using a batch size of 1 and an input size of 600 x 1000. Our design achieves inference times comparable to one-stage methods with the same backbone, showcasing its efficiency.
SGTR+ improves inference time by reducing the number of predicate node generator decoder layers, resulting in faster processing (\textbf{0.35s} to \textbf{0.22s} per image). Additionally, SGTR+ employs the lighter ResNet-50 backbone, reducing inference time by \textbf{48\%} (from 0.35s to 0.18s) while maintaining performance on par with SGTR with ResNet-101.

\noindent\textbf{Comapre with SOTA One-stage SGG}
{
Recently, many one-stage SGG methods are proposed by community.
Such as transformer based RelTR~\cite{cong2023reltr}, ISG~\cite{khandelwal2022iterative} and non-transformer based Structural Sparse RCNN (SSRCNN)~\cite{teng2021structured}. 
We detailed discuss the comparison between those SGG method with SGTR+.
}

\noindent$\bullet$
RelTR: We reproduced the RelRT by releasing code on the same hardware platform as the SGTR+ based on the same ResNet-50 backbone, and the results demonstrate that the SGTR+ achieves better performance and comparable time complexity.

\noindent$\bullet$
ISG: We compare SGTR+ with ISG, both under the same longtail strategy and without it. The results show that SGTR+ outperforms ISG \textbf{3.0} on mR@100, and \textbf{1.6} on hR@100.

\noindent$\bullet$
SSRCNN: Compared to SSRCNN, our SGTR+ model has significant advantages in terms of inference time and parameter complexity.
As shown in Fig.\ref{fig:ssrcnn-cmp}, the SGTR model has only half the inference time and \textbf{30\%} of the parameters of SSRCNN, which demonstrates that our model has significant advantages in terms of lightweight design while achieving comparable SGG performance.
The most time-consuming part of the SGTR+ is the self-attention module, which has an $O(N^2)$ time complexity that limits the inference speed.
The research community has proposed several efficient transformer designs \cite{wang2020linformer,zhu2020deformable, fan2021multiscale} that could further speed up our SGTR model.

Furthermore, we investigate why our model outperforms SSRCNN in body and tail categories but not head categories.
In contrast to SSRCNN's multi-scale ResNeXt-101 FPN backbone, our SGTR+ model's single-scale ResNet-101 backbone prevents head category relations retrieval, limiting entity detection performance.
Furthermore, it is non-trivial to effectively integrate the multi-scale feature due to the high complexity of self-attention, which is an interesting problem for future research.

\begin{table}
    \begin{center}
        \resizebox{0.45\textwidth}{!}{      
            \begin{tabular}{l|cccc|c}
                \toprule
                \textbf{Model}  & \textbf{mR@50}& \textbf{mR@100} & \textbf{R@50} & \textbf{R@100} & Det AP50  \\ \midrule
                BGNN  & 1.9 & 2.3 & 14.6 & 17.6 & \multirow{2}{*}{6.1} \\
                RelDN & 0.5 & 0.6 & 17.7 & 20.9 &  \\ \midrule
                SGTR  & 4.2 & 5.0 & 14.4 & 16.4 &   \multirow{3}{*}{6.0} \\
                SGTR+ & 4.2 & 5.7 & 15.2 & 17.4 &  \\
                SGTR+$^*$ & 8.4 & \textbf{10.3} & 8.4 & 10.2 &  \\
                \bottomrule
                \end{tabular}
            }
    \end{center}
    \vspace{-0.3cm}
    \caption{\textbf{The SGDet performance on GQA test-dev.} }
    \vspace{-0.68cm}
    \label{tab:gqa_performance} 
\end{table}

\vspace{1mm}
\noindent\textbf{GQA}~ 
Besides two typical SGG benchmarks, we also test our methods on a widely used cross-modal reasoning benchmark, GQA, as shown in Tab.~\ref{tab:gqa_performance}.
The initial motivation of SGG is to provide the structural representation for downstream visual reasoning tasks.
However, previous SGG work rarely tested their method on this dataset.
We evaluate the two-stage SGG baseline BGNN and our newly proposed SGTR family.
Our end-to-end design achieves an advantage in mean recall of \textbf{7.0} on a balanced metric, mR@100.
Since the GQA is a large-scale scene graph generation benchmark with 1703 entity categories and 311 predicate categories, it is more challenging than the other existing SGG benchmarks.

\vspace{-0.26cm}
\subsection{Qualitative Results}
\vspace{-0.1cm}
We demonstrate the qualitative analysis of SGTR+ in order to comprehend the mechanism of our newly proposed design more thoroughly.
The supplementary contains more qualitative results (prediction comparison, attention visualization, error mode analysis).

\noindent\textbf{Effectiveness of Spatial Constrained Predicate Node Generator}
As depicted in Fig.~\ref{fig:lyr_pred_cmp}, we visualize the prediction of each decoder layer from the predicate node generator to gain a better understanding of how spatial constraint design improves relation modeling.
In the first row of the SGTR+ intermediate prediction, explicit dynamic anchor generation generates high-quality relationship proposals from existing entity nodes.
The spatial constraint entity indicator sub-decoder improves entity and predicate representation and models relations using shallow decoders.
In contrast, in the lower part of Fig.~\ref{fig:lyr_pred_cmp}, there is a lot of instability in the relationship detection in the SGTR reasoning process. At the same time, since the lack of end-to-end optimization, the process of entity grouping is also disturbed by low-quality entity indicators.
{
Furthermore, in the right portion of Fig.~\ref{fig:lyr_pred_cmp}, we observe that the spatial-aware predicate decoder in SGTR+ effectively prioritizes essential features for predicate recognition, including entities and their interactions.
}

%% file: sec/conclu.tex
\vspace{-0.3cm}
\section{Conclusions}
We have introduced an end-to-end CNN-Transformer-based scene graph generating framework, termed as SGTR, and its extension SGTR+. 
Our framework consists of an entity-aware predicate representation for exploiting the compositional properties of visual relationships, and a bipartite graph construction process that inherits the advantages of both two-stage and one-stage methods. To improve the model efficiency, we further enhance the entity-predicate association modeling by redesigning the main module of the predicate node generator and graph assembly.
Our method achieves the state-of-the-art or competitive performance on the Visual Genome, Openimage V6, and GQA datasets across all metrics and with more efficient inference.

%% file: sec/supply.tex
\section*{Overview of Appendixes}
In this supplementary material, we present implementation details and more experiments results.
First, more experiments and analysis (\textit{e.g.}, analysis of overall recall performance, experiments using stronger long-tail learning strategy, unified graph assembling, and extra qualitative results) are described in Sec.~\ref{sec:more_exp}.
We provides the implementation details in Sec.~\ref{sec:implementation_detail}.
Moreover, we also present the details of graph assembling mechanism and loss function in Sec.~\ref{sec:assembel} and Sec.~\ref{sec:loss_equations}.

In this supplementary material, we present implementation details and more experimental results.
First, more experiments and analysis (\textit{e.g.}, analysis of overall recall performance, experiments using stronger long-tail learning strategies, and extra qualitative results) are described in Sec.~\ref{sec:more_exp}.
We provide the implementation details in Sec.~\ref{sec:implementation_detail}.
Moreover, we also present the details of the graph assembling mechanism and loss function in Sec.~\ref{sec:assembel} and Sec.~\ref{sec:loss_equations}.

\section{More Experimental Results}\label{sec:more_exp}
\subsection{Overall Recall Analysis}

\begin{figure}[h]
    \centering
    \includegraphics[width=\linewidth]{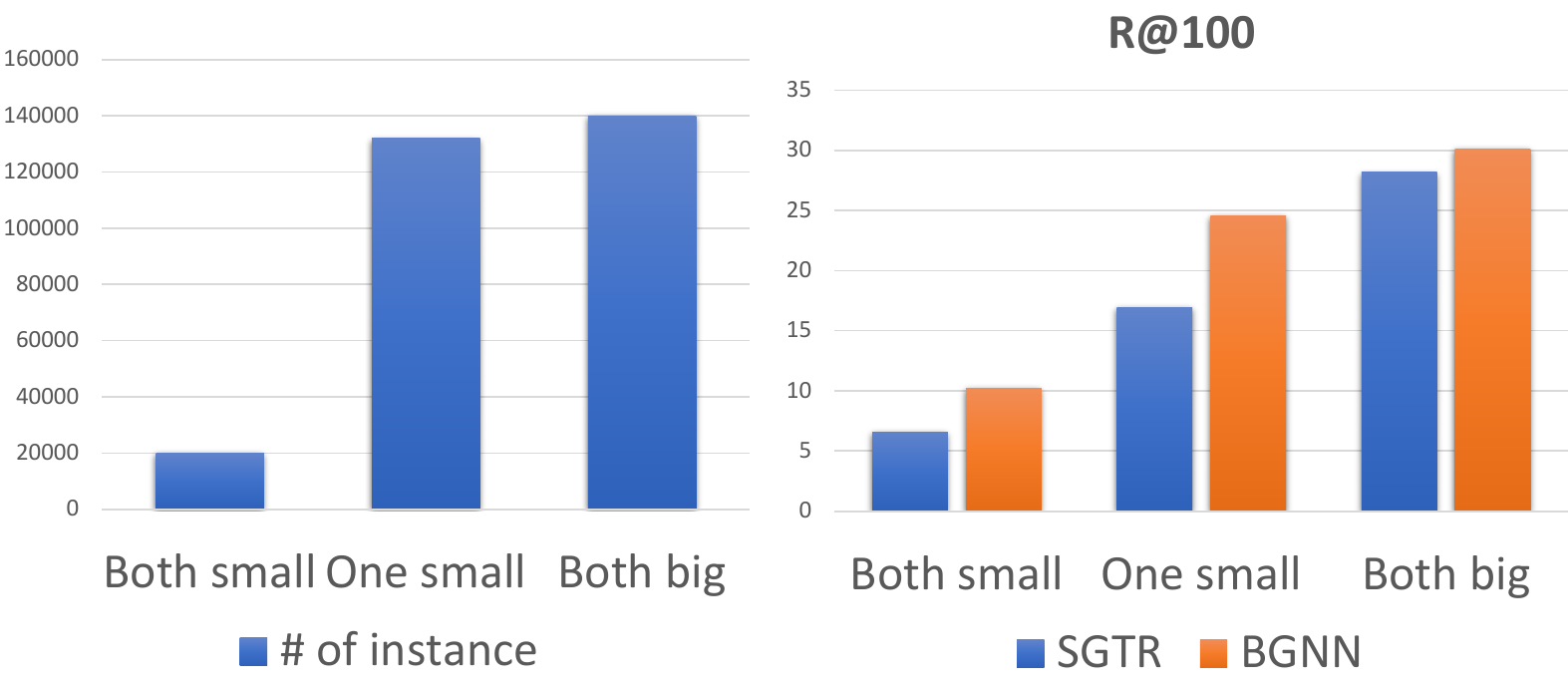}
    \caption{\textbf{Left)} The statistics of entity sizes in visual relationship based on the training set of Visual Genome. 
    \textbf{Right)} The performance (Recall@100) of relationship detection with respect to different entity sizes.
    We define an entity as "small" if the area of its box is smaller than 64 $\times$ 64 pixels.
    }\label{fig:rel_stat}
\end{figure}

First, we analyze why our SGTR achieves lower overall recall performance than the traditional two-stage design.
One potential reason is that the SGTR uses an ResNet-101 backbone (like DETR) for entity detection rather than the Faster-RCNN with an ResNet-101 FPN backbone.
According to the experimental results of DETR~\cite{carion2020end}, using the ResNet-101 backbone achieves lower performance than Faster-RCNN with the ResNet-101 FPN on small objects (21.9 versus 27.2 APs on the COCO dataset).

To confirm that, we further study how small entities influence visual relationship detection.
We categorize the relationship instances in the Visual Genome dataset into three disjoint sets according to their entity sizes, and plot the statistics of the sets in Fig.~\ref{fig:rel_stat}. 
The result shows that more than half of the relationships consist of small entities. 
We also compare the performance of our method (R@100) on three relationship sub-sets with the two-stage approach BGNN~\cite{li2021bipartite} and report the performance in Fig.~\ref{fig:rel_stat}.
The BGNN with the two-stage design outperforms the SGTR on the relationships with small entities by a large margin due to the inherent limitation of DETR on small entity detection.

The issue of recognizing small objects efficiently and effectively is still under active study~\cite{yao2021efficient, wang2021anchor, zhu2020deformable, meng2021conditional}.
With more sophisticated transformer-based detectors, the overall recall of our method can be further improved.

\subsection{Influence of Object Detector} 

We first compare the performances of DETR and faster RCNN on the head (H), body (B) and tail (T) predicates in Tab.\ref{tab:ent_part}, which also groups the detector results into H/B/T object classes. The result indicates that DETR achieves a similar detection performance as Faster R-CNN for tail predicates. Moreover, we equip BGNN with DETR and as shown in Tab.\ref{tab:ent_val}, its performance is on par with its Faster R-CNN counterparts.
Finally, we also combine DETR optimized in SGTR with a finetuned BGNN. The results in Tab.\ref{tab:ent_val} clearly show the SGTR outperforms this baseline, demonstrating the benefit of our design.

\begin{table}[h!]
    \centering
        \resizebox{0.38\textwidth}{!}{        
            \begin{tabular}{l|ccc}
                \toprule
            & \multicolumn{3}{c}{\textbf{Recall} (S-DETR / RCNN)}  \\
            & Rel Head & Rel Body & Rel Tail  \\ \midrule
                Ent Head &  49.3 / \textbf{50.5}   &  49.2 / \textbf{50.3} & \textbf{59.7} / 57.1   \\
                Ent Body &  39.5 / \textbf{41.3}   &  \textbf{46.2} / 43.1 & \textbf{44.9} / 43.7   \\
                Ent Tail &  \textbf{44.3} / 41.7  &   \textbf{47.6} / 44.5 & \textbf{45.1} / 38.5    \\ 
                \bottomrule
            \end{tabular}
        }
        \caption{The entity detection performances of DETR and faster RCNN on the head (H), body (B) and tail (T) predicates.} 
        \label{tab:ent_part}
 \end{table}

\begin{table}[h!]
    \centering
        \resizebox{0.47\textwidth}{!}{        
            \begin{tabular}{l|cc|ccc|cc}
                \toprule
                M & \textbf{mR@50/100} & \textbf{R@50/100} & \textbf{H}  & \textbf{B}  & \textbf{T} & \textbf{AP}@det & \textbf{mR}@det \\ \midrule
                BGNN  &   8.4 / 9.8  &   \textbf{29.0} / \textbf{33.7}    &  \textbf{30.0} & 11.2  & 2.2   & 29.9 & 41.1 \\ 
                BGNN$_\text{D}$  &  6.3 / 7.7   &   25.4 / 29.5   & 28.5  & 7.2  & 1.5 & 30.7 & \textbf{41.2} \\ 
                BGNN$_\text{S}$  &  6.7 / 8.3   &  25.6 / 29.7    & 28.5  & 7.2  &  3.0 & \textbf{31.2} & 41.0 \\
                SGTR$^\dagger$  &  \textbf{14.4} / \textbf{17.7}   &  24.8 / 28.5      & 21.7  & \textbf{21.6} & \textbf{17.1}  & \textbf{31.2}  & 41.0 \\ \bottomrule
                \end{tabular}
        }
        \caption{ 'D' means DETR. 'S' means DETR optimized in SGTR.
        $\dagger$ means bi-level resampling, 'AP': mAP50, and 'mR': mRecall} 
        \label{tab:ent_val} 

 \end{table}

\subsection{Experiments with Long-tail Learning Strategy}

\begin{table}[!ht]
    \begin{center}
        \resizebox{\linewidth}{!}{        
                \begin{tabular}{l|cc|ccc}
                \toprule
                  \textbf{Method}       & \textbf{mR@50/100} & \textbf{R@50/100} & \textbf{Head}  & \textbf{Body}  & \textbf{Tail}  \\ \midrule
                 \textbf{Ours}        & 12.0~/~15.2  & \textbf{24.6}~/~\textbf{28.4}& \textbf{28.2} & 18.6 & 7.1      \\
                 \textbf{Ours}$^{*}$  & \textbf{15.8}~/~\textbf{20.1} & 20.6~/~25.0 & 21.7 & \textbf{21.6} & \textbf{17.1}   \\
                 \textbf{Ours}-P &  18.9 / 22.0   &   \textbf{22.1} / \textbf{24.8}   & \textbf{26.0}  & 20.9  & \textbf{15.2}  \\  \midrule
                 \textbf{Ours}$^{\text{DisAlign}}$  & 13.7~/~16.8 & \textbf{24.1}~/~\textbf{28.0} & \textbf{26.8} & {21.7} & {8.9}   \\
                 \textbf{Ours}$^{\text{ACBS}}$  & 16.5~/~19.8 & 20.8~/~23.6 & 23.4 & 21.6 & 17.5   \\ 
                \textbf{Ours}$^{\text{cRT}}$  & \textbf{18.8}~/~\textbf{21.6} & {22.0}~/~{24.8} & {24.1} & \textbf{22.1} & \textbf{18.1}   \\
                  \bottomrule
                \end{tabular}
        }
    \end{center}
    \caption{\textbf{The performance of SGTR by adopting the advanced long-tail learning strategy on the VG dataset.}
    "*" denotes the bi-level sampling proposed in \cite{li2021bipartite};
    "P" denotes the modified bi-level sampling;
    The "${\text{cRT}}$" denotes the decoupled retraining strategy on predicate classifier proposed by~\cite{kang2019decoupling}; the "DisAlign" denotes the retraining strategy for logits adjustment proposed by~\cite{Zhang_2021_CVPR}; 
    the "${\text{ACBS}}$" refers to the alternative class balanced retraining strategy proposed by~\cite{desai2021learning}. } 
    \label{tab:train_stragt}
\end{table}

Long-tail data distribution is a challenging issue in the SGG.
To achieve better performance on the SGG task, we further apply several recent long-tail learning strategies in our model.~\cite{kang2019decoupling, Zhang_2021_CVPR, desai2021learning}, and report the performance in Tab.~\ref{tab:train_stragt}. We find that there exists a trade-off between overall and mean recall in the comprehensive experimental results.
The advanced learning strategies enable our model to either achieve a higher mean recall or maintain a better trade-off between overall and mean recall.

\begin{figure}
    \centering
    \includegraphics[width=1.0\linewidth]{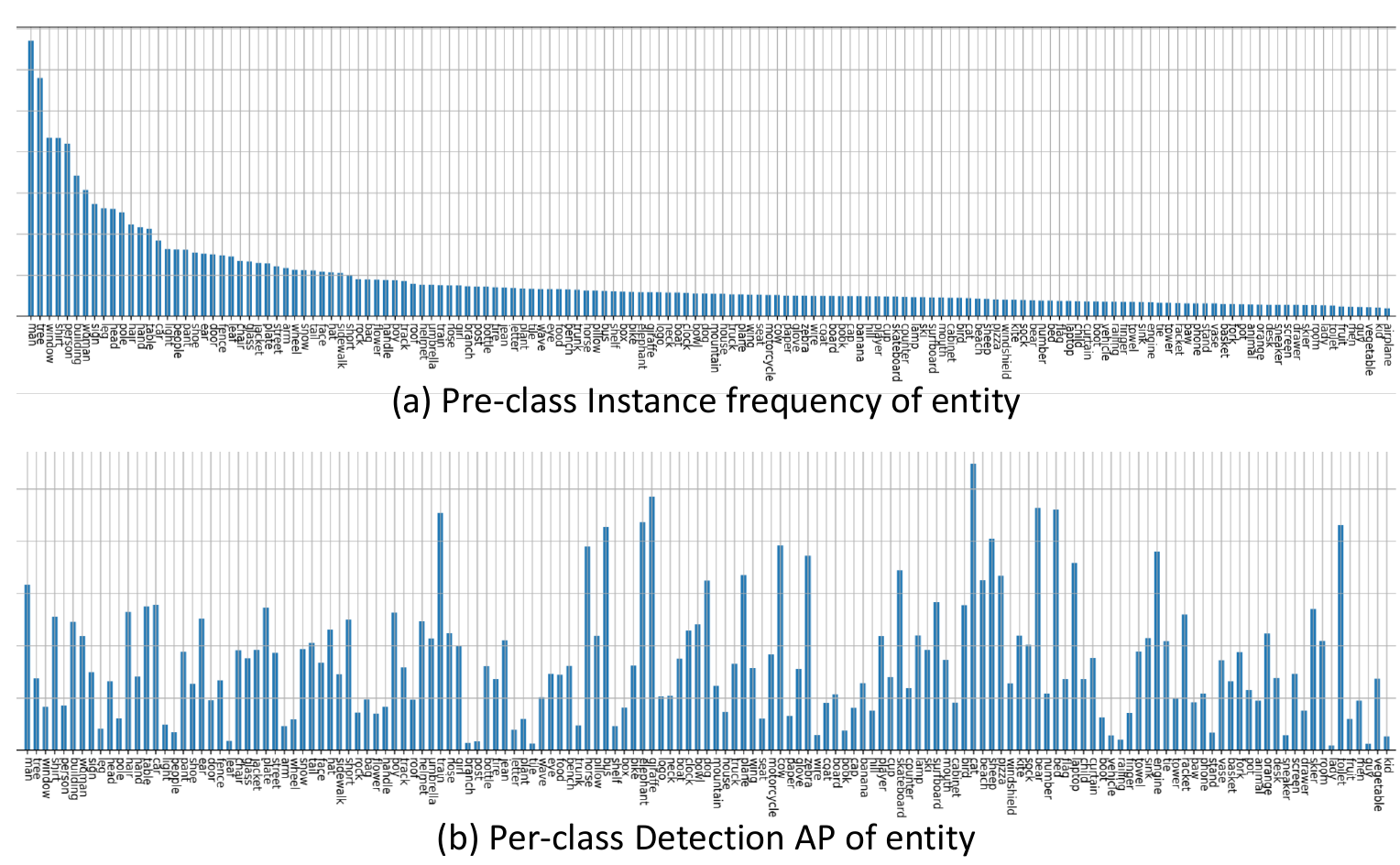}
    \caption{\textbf{The frequency statistics and per-class detection performance of entities on the VG dataset}. \textbf{(a)}: The frequency of each entity category; \textbf{(b)} The per-class entity detection performance (AP).
    } \label{fig:ent_cls_feq}
\end{figure}

\noindent$\bullet$ We adopt the retraining strategy, DisAlign~\cite{Zhang_2021_CVPR}, to adjust the predicate prediction logits via loss re-weighting with respect to the instance distribution of relationships. 
This method improves the performance trade-off between mR@100 and R@100 by increasing mean recall by \textbf{1.6} and achieving only a 0.4 performance drop on R@100.

\noindent$\bullet$ Moreover, we re-implement the alternative class-balanced retraining strategy (ACBS)~\cite{desai2021learning}, which achieves the SOTA on mean recall with the two-stage SGG model.
The ACBS retrains both entity and predicate classifiers using class-based sampling. This method achieves high mean recall performance while sacrificing the performance of the overall performance.


\noindent$\bullet$ Finally, we apply the decoupled retraining strategy~\cite{kang2019decoupling} on the predicate classifier of SGTR.
We observe that using additional balanced-sampling retraining results in \textbf{6.4} performance gain on mR@100 with a 3.6 drop on R@100. This strategy outperforms the aforementioned methods in terms of mean recall performance.

\noindent$\bullet$ 
We also observe that \textit{using the re-balance idea on entity classifier does not bring too much performance benefit.}  To investigate this phenomenon, we report the relationship between the per-class instance frequency and the performance of entity detection in Fig.~\ref{fig:ent_cls_feq}.
Despite the fact that the distribution of entity instances obeys the long-tail distribution, SGTR's entity detection performance is quite balanced, which means that the transformer-based detector is capable of tackling the data imbalanced scenario to some degree, and the additional re-balancing strategy is unnecessary.

 \subsection{Model Selection} 
 We present experiments for selecting the hyper-parameters ($N_r$, $K$) on validation set of VG in Tab.\ref{tab:num_q}. The performance saturates at $N_r=160$ and $K=3$ .
 \begin{table}[h!]
    \centering
        \resizebox{0.36\textwidth}{!}{        
            \begin{tabular}{l|cc|l|cc}
                \toprule
                $N_r$ & \textbf{mR@100} &  \textbf{R@100} & $K$ & \textbf{mR@100} &  \textbf{R@100}  \\ \midrule
                100   &  16.1  &   27.5 & 1 & 16.4 & 26.2\\
                130   &  16.3  &   27.3 & 2 & 17.3  & 28.2\\
                160   &  \textbf{17.7 }  &   \textbf{28.5} & 3  &  \textbf{17.7}  & \textbf{28.5} \\ 
                190   &  16.1  &   27.0 & 4 &  17.5  & 28.2\\ \bottomrule
                \end{tabular}
        }
        \caption{The performance of different choices of hyper-parameters $N_r$.}
        \label{tab:num_q} 
         
 \end{table}

\begin{figure}[h!]
    \centering
    \includegraphics[width=0.97\linewidth]{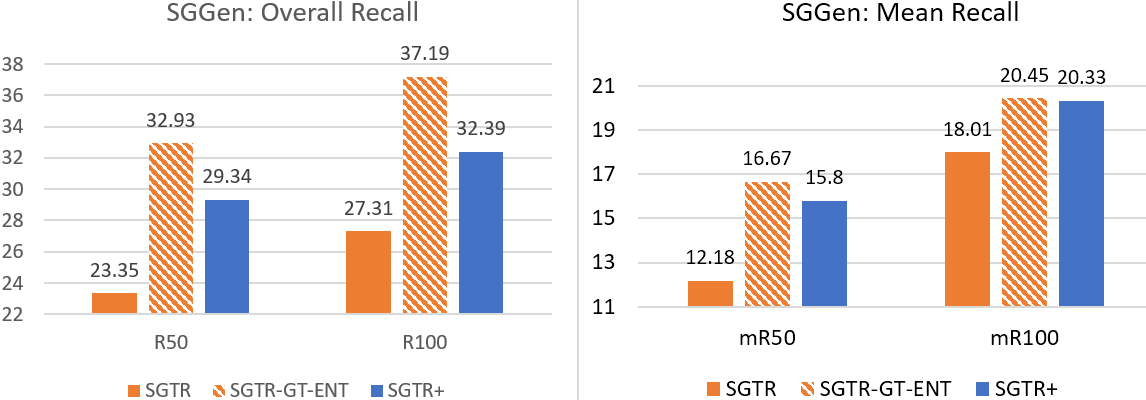}
    \caption{\textbf{The empirical study for upper bound of entity modeling.}
    The upper-bound uses GT entities for SGTR\cite{li2022sgtr}  on Visual Genome dataset.
    GT-ENT stands for using GT entities for predicate decoder.}
    \label{fig:emprical_sty}
\end{figure}

\subsection{Efficacy of Entity Modeling for SGTR Framework} 
To prove entity modelling works, we feed the predicate node generator the GT entity to decode the predicate and entity association.
Fig.~\ref{fig:emprical_sty} demonstrates a significant increase in R@100 from 27.31 to 37.19 and mR@100 from 18.01 to 20.45 through upper bound analysis.

\begin{table}
    \begin{center}
        \resizebox{0.49\textwidth}{!}{      
            \begin{tabular}{l|cccc|ccc}
                \toprule
                \textbf{Tau}  & \textbf{mR@50}& \textbf{mR@100} & \textbf{R@50} & \textbf{R@100} & Head & Body & Tail  \\ \midrule
                L:1e-0 & \textbf{16.6} & \textbf{22.4} & \textbf{26.3} & \textbf{30.8}  &  \textbf{29.4} &  \textbf{22.5} &  \textbf{19.1}  \\
                L:3e-3 & 13.4 & 19.8 & 26.1 & 30.4 & 29.5 & 23.1 & 13.6 \\
                F:1e-0 & 7.0  & 9.9 & 18.1 & 20.5 & 19.7 & 12.0 & 4.8  \\
                F:3e-3 & 13.5 & 20.0 & 26.0 & 30.0 & 29.3 & 22.9 & 14.1 \\
                \bottomrule
                \end{tabular}
            }
    \end{center}
    \caption{\textbf{Ablation study on $\tau$ of unified graph assembling.} L: indicates the $\tau$ is learnable during the training; F: indicates the $\tau$ is fixed in training.} 
    \label{tab:sgtr+uga_tau} 
\end{table}

\subsection{Parameter $\tau$ of Unified Graph Assembling}
\begin{figure}
    \centering
    \includegraphics[width=0.32\textwidth]{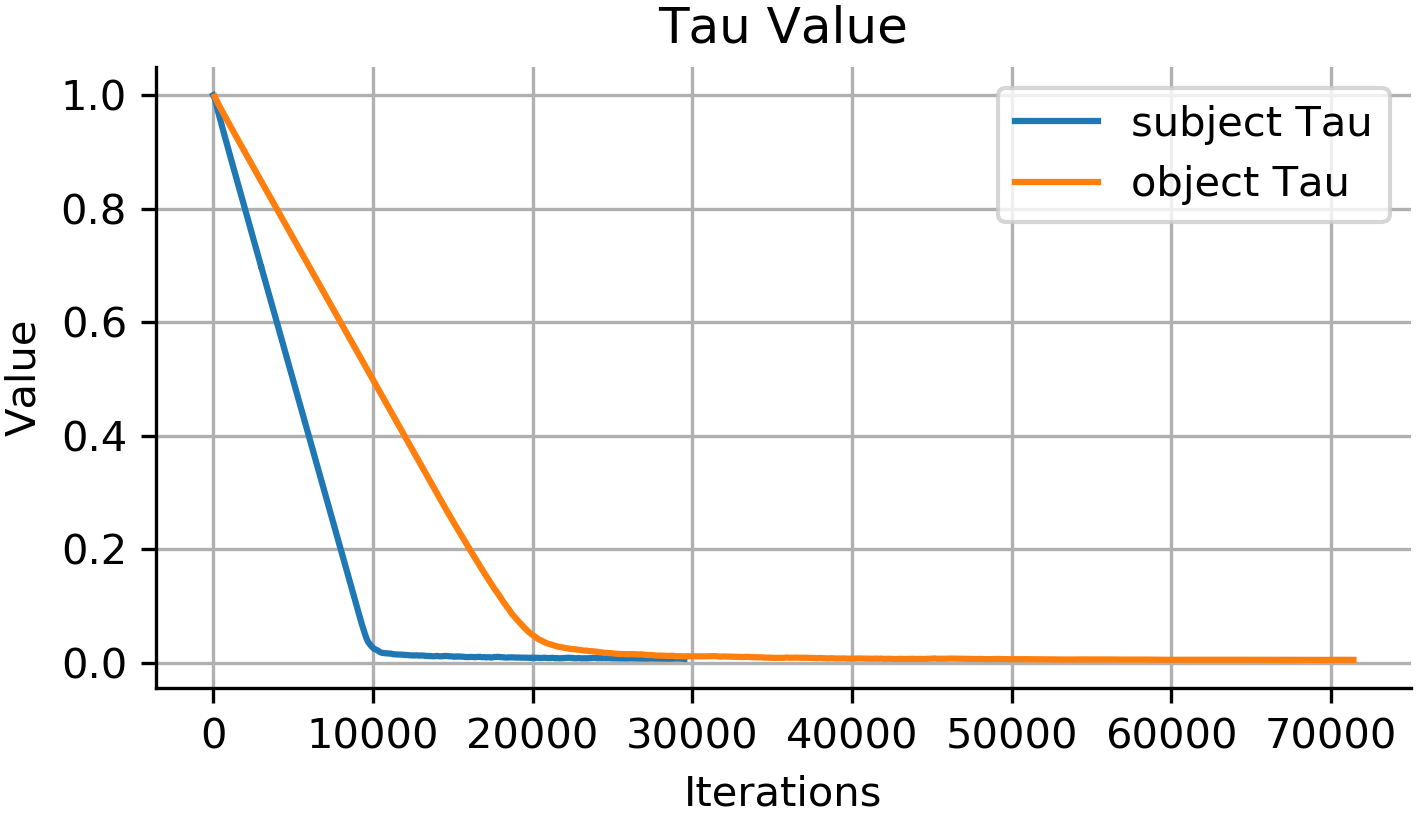}
    \caption{\textbf{The visualization of $\tau_{ga}$ of unified graph assembling during the training.}}
    \label{fig:uga_tau}
\end{figure}

The unified graph assembling introduces a parameter $\tau$ for calibrating softmax during the calculation of the similarity matrix between entity nodes and entity indicators.
We inspecting tau's tendency for subject and object entity assembling during training as shown in Fig. \ref{fig:uga_tau}.

\noindent$\bullet$ 
The similarity matrix's value distribution sharpens with  $\tau$.
During the train, higher  $\tau$ values in earlier stages allow smoother aggregation of entity nodes to construct triples and gradually decrease for more specific entity grouping since the quality of the entity indicator is getting better.
We evaluate the different choice for initialize the $\tau$ refer to supplementary.

\noindent$\bullet$ 
We examine how $\tau$ initialization affects model learning in Tab.\ref{tab:sgtr+uga_tau}.
As shown in line 3, if we disable the $\tau$ by setting it as constant 1, all relation categories have a large performance drop, demonstrating the need to recalibrate the softmax for similarity matrix.
By comparison with lines 4 and 1, the result indicates the model benefits from a dynamic adjustment of tau during training.
Learnable $\tau=1.0$ outperforms fixed or learnable with $3*10^{-3}$ initialization on all metrics.

\begin{figure*}[h]
    \centering
    \includegraphics[width=0.8\linewidth]{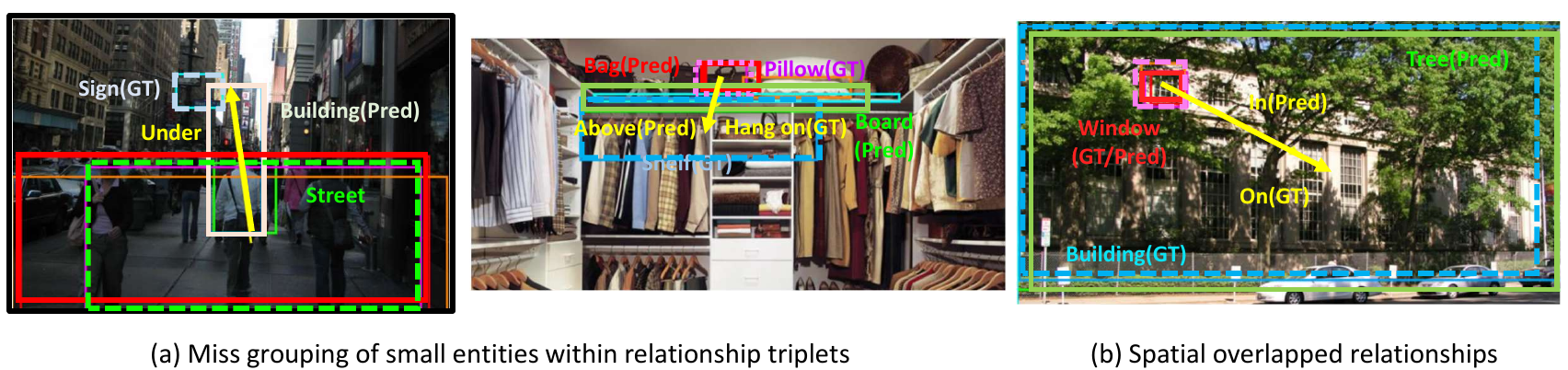}
    \caption{\textbf{The error-mode analysis of SGTR.} The figure shows the two primary error of SGTR. } 
    \label{fig:error-mode}
\end{figure*}

\subsection{Qualitative Results}

\begin{figure*}
    \centering
    \includegraphics[width=\textwidth]{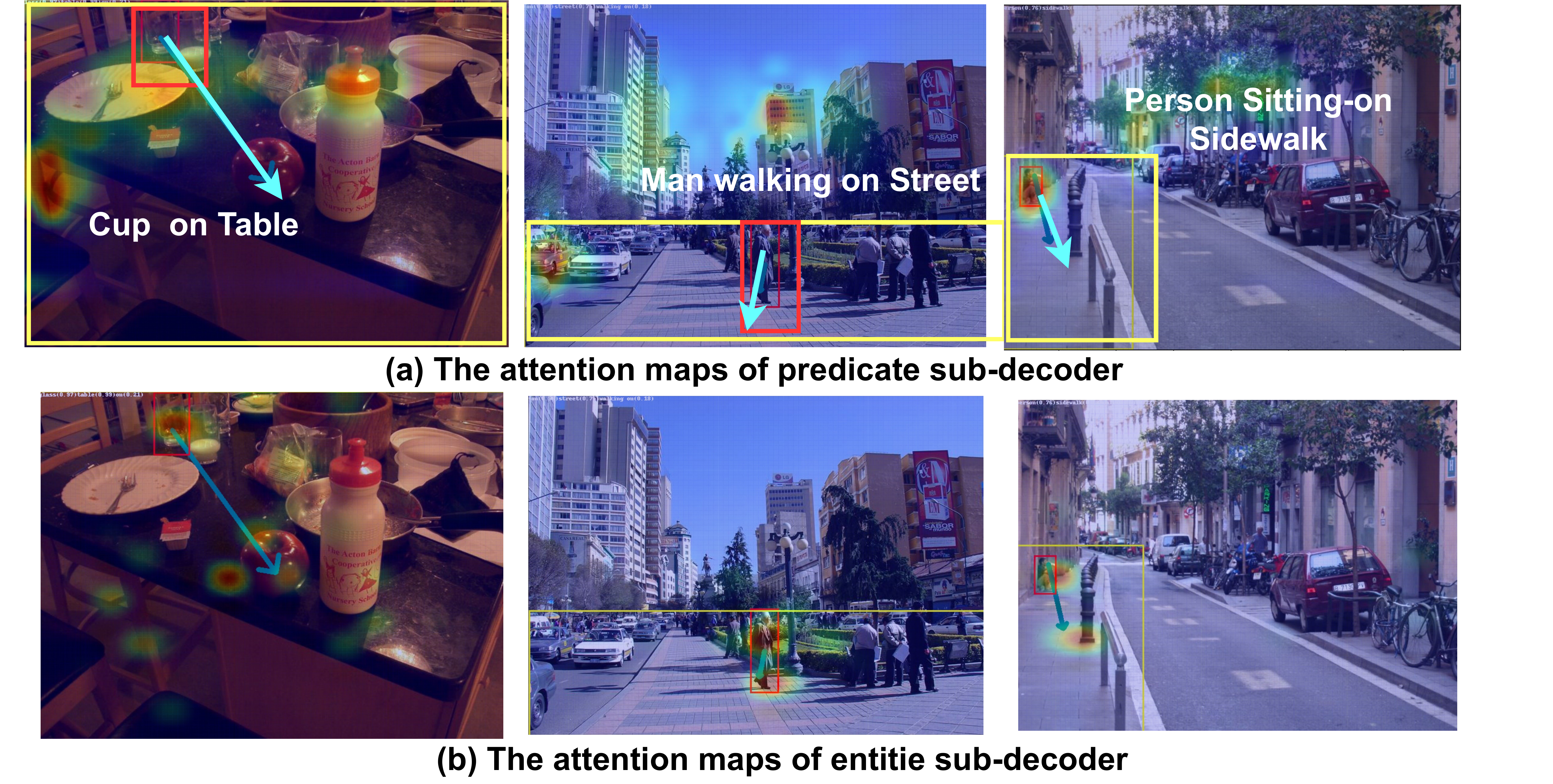}
    \caption{\textbf{The visualization on attention from structural predicate decoder of SGTR.} The predicate sub-decoder focus on contextual representation around the entities of triplets.
Entity indicator sub-decoders focus on relationship-based entity regions.}
    \label{fig:vis}
\end{figure*}

\noindent\textbf{Visualization of Decoders' Attention }
As shown in Fig. \ref{fig:vis}, we visualize the attention weight of the predicates sub-decoder and entity sub-decoder on images from the validation set of the Visual Genome dataset.
By comparing the heatmaps in Fig. \ref{fig:vis} (a) and Fig. \ref{fig:vis} (b),
we note that for the same triplet prediction, the predicate sub-decoder focuses more on the contextual regions around the entities of triplets while the entity sub-decoders put more attention on the entity regions.
Therefore, our design allows the model to learn the compositional property of visual relationships more effectively, which improves prediction accuracy.

\begin{figure*}
    \centering
    \includegraphics[width=\linewidth]{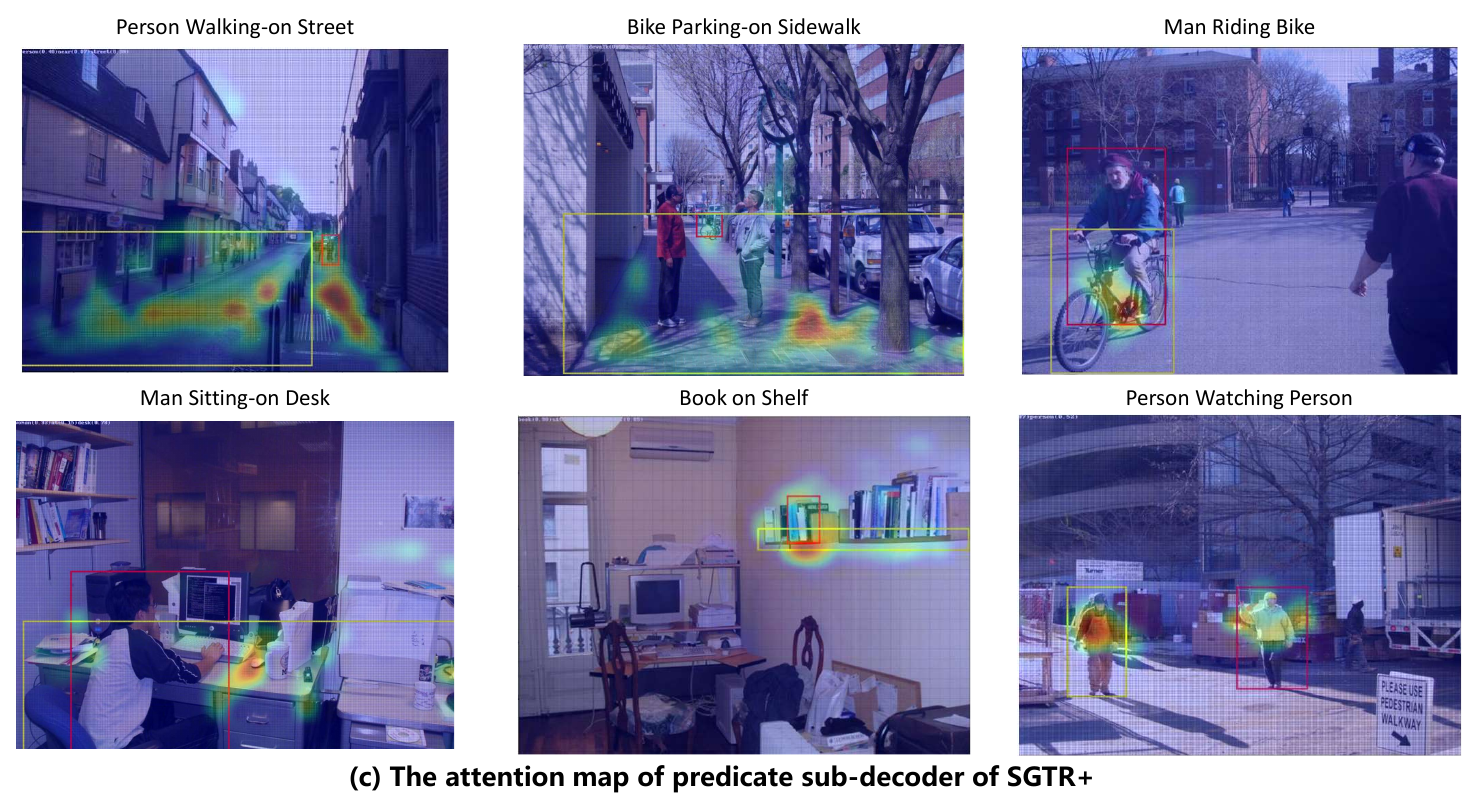}
    \caption{\textbf{The visualization on attention heatmap of predicate decoder of SGTR+.}  }
    \label{fig:vis_att_sgtr+}
\end{figure*}

We further visualize the cross-attention within the predicate sub-decoder of SGTR+. With integration of explicit spatial-aware decoding, leading to notable improvements in the interpretability of attention maps, as exemplified in Fig.~\ref{fig:vis_att_sgtr+}.
This visualization underscores that the decoder predominantly emphasizes the semantic aspects of both subject and object entities, aiming to extract discerning features essential for predicate recognition.

\noindent\textbf{Error Mode Analysis}
To more intuitively understand the weakness of our SGTR, we find the following typical error-mode by visualizing the predictions. 
As shown in Fig.~\ref{fig:error-mode}, there are two types of error-modes: miss-grouping the small entities and detecting the highly overlapped relationships.

\begin{figure*}[h]
    \centering
    \includegraphics[width=\linewidth]{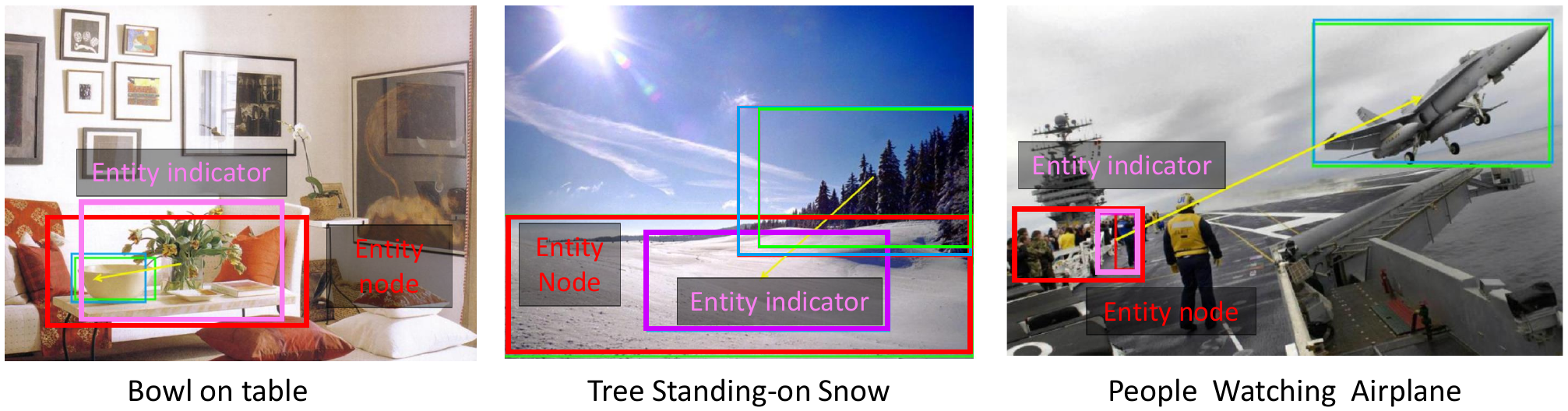}
    \caption{\textbf{Comparison of entity indicator of predicate node and entity node.} We use the different colored bounding boxes of entities to distinguish between the entity node (red) and the entity indicator (pink). The yellow arrow indicates the predicate between the entities. (best viewed in color)}
    \label{fig:rel_dec}
\end{figure*}

\noindent\textbf{Visualization of Entity Indicator and Entity Node}
To demonstrate the effectiveness of our proposed graph assembling mechanism, we visualize predictions of entity indicators of our predicate representation and entity nodes after the graph assembling.
As shown in Fig.~\ref{fig:rel_dec}, the entity indicator only provides a rough localization and classification of entities rather than precise bounding boxes.
This information can be refined into more accurate entity results with graph assembling, which significantly improves the quality of the generated scene graph.

\begin{figure*}[h]
    \centering
    \includegraphics[width=\linewidth]{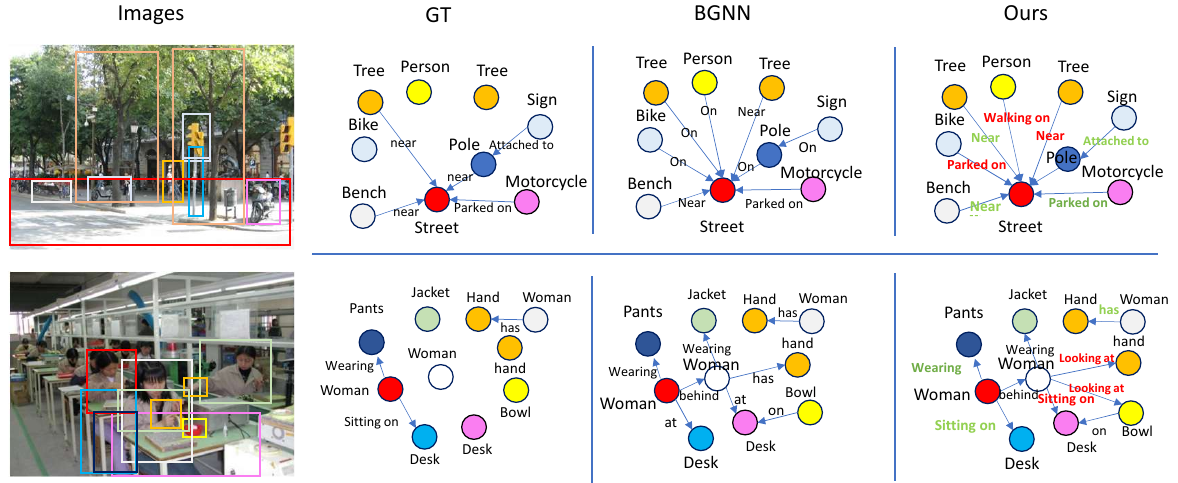}
    \caption{\textbf{Qualitative comparison between our method and BGNN$\dagger$ in SGG.} Both methods predict many reasonable relationships which are not annotated in GT. We mark the relationships of rare semantic predicate categories retrieved by SGTR+ in \textcolor{red}{\textbf{red}}, and prediction matched GT in \textcolor{green}{\textbf{green}} (best viewed in color). }
    \label{fig:vis_res}
\end{figure*}

\noindent\textbf{Prediction Comparison between Different Design}
We compare different method~(\textit{e.g.}, BGNN~\cite{li2021bipartite}) by visualizing the relationship predictions.
In Fig. \ref{fig:vis_res}, we mark the different relationship predictions between BGNN and SGTR with \textcolor{red}{red} color.
It shows that SGTR retrieves more relationships of \textit{less frequent semantic categories} than BGNN.

\section{Implementation Details}\label{sec:implementation_detail}
We implement our method based on the PyTorch 1.8~\cite{NEURIPS2019_9015} and cvpods~\cite{zhu2020cvpods}.
Our training process consists of two phases: \textit{ 1) entity detector pre-training} and \textit{2) SGTR joint training}.

\noindent\textbf{Entity Detector Pre-training Phase:} 
We follow the DETR training configuration to learn the entity detector on Visual Genome and Openimage datasets.
We train the entity detector with the AdamW optimizer with a learning rate of 1e-5, a batch size of 16, and the model takes 100 epochs for convergence on 4 TITAN V GPUs.
We use the same scale augmentation with DETR, resizing the input images such that the shortest side is at least 480 and at most 600 pixels, while the longest is at most 1000.
The hyper-parameters of Transformer (\textit{e.g.,} number of attention heads, drop-out rate) are also kept the same with the DETR.

\noindent\textbf{Joint Learning Phase:} 
In the joint training phase, we adopt the same optimizer, learning rate, and batch size configuration as in the entity detector pre-training stage.
In contrast with the two-stage SGG model, we refine the parameters of the detector in the joint learning rather than freezing the detector. 
We empirically observe that this refinement further improves the performance of entity detection.
We train the SGTR for the Visual Genome dataset for 8.39e4 iterations without learning rate decay by using an early-stopping strategy. 
For the Openimage dataset, we train the model with 1.5e5 iterations, and the learning rate is decreased by 0.1x after 1e5 iterations.

\noindent\textbf{Entity Detector Setup}
The input of SGTR and SGTR+ for DETR based entity node generator shares same pre-proces of previous two-stage SGG design. We resize the shortest edge to 600 and max size to 1000. The DETR setup is also keep the same setup with the 100 query proposals. The detected entity nodes are directly input for the following modules.

\noindent\textbf{Two-stage SGG Methods Setup}
We re-implement the previous two-stage methods (BGNN~\cite{li2021bipartite} and RelDN~\cite{zhang_relpn_2017}) in our codebase by using the same configuration as the released codes. For fair comprasion, we replace the ResNeXt-101 FPN backbone with the ResNet-101 backbone to learn the model.
We also apply the one-stage HOI works~(AS-Net~\cite{chen2021reformulating} and HOTR~\cite{kim2021hotr}) on the SGG.
We use the hyper-parameters of the models reported by the authors. All experiments are conducted by training the model until convergence.

\section{Correspondence Matrix for Assembling}\label{sec:assembel} 

For clarity, we will use the correspondence matrix between the subject entity and predicate $\mathbf{M}_s$ to introduce the details.
The correspondence matrix is determined by the distance function, which takes the semantic outputs~(\textit{e.g.} bounding boxes $\mathbf{B}$, classification $\mathbf{P}$  of the entity detector and entity indicator of the predicate structural decoder) as input.
Specifically, the distance function consists of two parts: spatial matching distance $d_{loc} \in \mathbb{R}^{N_r \times N_e}$ and category matching distance $d_{cls} \in \mathbb{R}^{N_r \times N_e}$, as shown in Eq.~\ref{eq:m_s}
\begin{align}
    \mathbf{M}^s &=d_{loc}(\mathbf{B}_s,\mathbf{B}_e)\cdot d_{cls}(\mathbf{P}_s,\mathbf{P}_e) \label{eq:m_s}
\end{align}
Each element in the correspondence matrix $\mathbf{M}^s$ is calculated by pairing the $N_r$ predicate predictions with $N_e$ entity predictions, as shown in the following equations:
\begin{align}
    \mathbf{M}^s_{i, j} &= d_{loc}(\mathbf{B}_s(i),\mathbf{B}_e(j))\cdot d_{cls}(\mathbf{P}_s(i),\mathbf{P}_e(j)) \\
    &= d_{loc}(\mathbf{b}_{s, i},\mathbf{b}_{e, j})\cdot d_{cls}(\mathbf{p}_{s, i},\mathbf{p}_{e, j}) 
\end{align}
where $i \in [0, N_e], j \in [0,N_r]$ for enumerating each pair between the predicate proposal and entity set.

Then we present the two components of the distance function, $d_{loc}$ and $d_{cls}$.
Specifically, the $d_{loc}$ consists of the $d_{giou} \in \mathbb{R}^{N_r \times N_e}$ and $d_{center} \in \mathbb{R}^{N_r \times N_e}$, as show in Eq.~ \ref{eq:d_loc}.
\begin{align}
    d_{loc}(\mathbf{b}_s,\mathbf{b}_e) &= \frac{d_{giou}(\mathbf{b}_s,\mathbf{b}_e)} {d_{center}(\mathbf{b}_s,\mathbf{b}_e)} \label{eq:d_loc}
\end{align}
Concretely, the $d_{giou}$ is the clipped GIOU of the entity and the indicator's bounding boxes, and $d_{center}$ is the L1 distance between  the bounding boxes' centers in Eq. \ref{eq:d_giou}, \ref{eq:d_center}.
The center points-based matching has also been adopted in HOI methods~\cite{chen2021reformulating, liao2020ppdm, wang2020learning}.
\begin{align}
     d_{giou}(\mathbf{b}_s,\mathbf{b}_e) &= \max(\min(\text{GIOU}(\mathbf{b}_s,\mathbf{b}_e), 0), 1) \label{eq:d_giou}\\
     d_{center}(\mathbf{b}_s,\mathbf{b}_e) &= || [x_c,y_c]^s_i - [x_e,y_e]^e_i ||_1 \label{eq:d_center}
\end{align}
Here $[x_s,y_s]^s$ and $[x_e,y_e]^e$ are the normalized center coordinates of the bounding box in $\mathbf{b}_s$ and $\mathbf{b}_e$ respectively.
For the $d_{cls}$, we use the cosine distance to calculate the similarity of the classification distribution between two entity predictions, as shown in following equation:
\begin{align}
     d_{cls}(\mathbf{p}_s, \mathbf{p}_e) = \frac{\mathbf{p}_s \cdot
     \mathbf{p}_e^\intercal}{||\mathbf{p}_s|| \cdot ||\mathbf{p}_e||}  
\end{align}

\section{Matching Cost and Loss Function}\label{sec:loss_equations}
\subsection{Triplets Matching Cost}

We use the set-matching strategy to supervise the relationship predictions~$\mathcal{T}=\{(\mathbf{b}_e^s,\mathbf{p}_e^s, \mathbf{b}_{e}^o,\mathbf{p}_e^o, \mathbf{p}_p,\mathbf{b}_p)\}$.
To obtain the matches, we need to calculate and minimize the matching cost $\mathcal{C} \in \mathbb{R}^{N_r \times N_{gt}}$ between the $N_r$ relationship predictions and the $N_{gt}$ GT relationships. 
Concretely, the matching cost $\mathcal{C}$ includes two parts: the predicate cost $\mathcal{C}_p$ and the entity cost $\mathcal{C}_e$, as: 
\begin{align}
    \mathcal{C} = \lambda_{p} \mathcal{C}_p + \lambda_{e} \mathcal{C}_e 
\end{align}
where $\lambda_{p}, \lambda_{e}$ is the coefficients of two cost terms.

The predicate cost, $\mathcal{C}_p(i, j)$ between the $i$-th predicate prediction and the $j$-th ground-truth relationship is computed according to the predicate classification distribution and location prediction in Eq.~\ref{eq:c_p}: 
\begin{align}
    \mathcal{C}_p (i, j) 
     = & \exp \left( - \mathbf{p}^{gt}_{p,j} \cdot \mathbf{p}_{p,i}^\intercal \right)   + \Vert \mathbf{b}_{p,i}  - \mathbf{b}_{p,j}^{gt} \Vert_1 \label{eq:c_p}
\end{align}
where $\mathbf{p}_{p,i}  \in \mathbb{R}^{1 \times C_p} $ is the $i$-th $\mathbf{P}_{p}$, and $\mathbf{p}^{gt}_{p,j} \in \mathbb{R}^{ 1 \times C_p}$ is the one-hot predicate label of the $j$-th ground truth relationship. 
Similarly, $\mathbf{b}_{p,i} \in \mathbb{R}^{1 \times 4}$ and $\mathbf{b}_{p,j}^{gt} \in \mathbb{R}^{1 \times 4}$ are the center coordinates of the entity pair from the $i$-th relationship prediction and the $j$-th ground truth relationship, respectively.

The entity cost $\mathcal{C}_e(i, j)$ between the $i$-th predicted relationship and $j$-th ground-truth relationship is given by:
\begin{align}
    \mathcal{C}_e(i, j)  =  
    & w_{giou} \cdot \prod_{\star=\{s, o\}} \text{exp} \left( -
        d_{giou}(\mathbf{b}^\star_{e,i},\mathbf{b}^\star_{gt,j}) \right) \\
    &+ w_{l1} \cdot \sum_{\star=\{s, o\}} || \mathbf{{b}}^\star_{e,i} -\mathbf{b}^\star_{gt,j} ||_1 \\ 
    & + w_{cls} \cdot \prod_{\star=\{s, o\}} \exp \left( - \mathbf{p}_{e,j}^{(\star ,gt)} \cdot {\mathbf{p}^\star_{e,i}}^\intercal \right)
\end{align}
where the $\mathbf{b}^\star_{e,i}$ and $\mathbf{p}^\star_{e,i}$ are the $i$-th subject/object entity box and category distribution of relationship $\mathcal{T}$ after graph assembling, respectively.
The $\mathbf{b}^\star_{gt,j}$ and $\mathbf{p}_{e,j}^{(\star,gt)}$ is the subject/object bounding boxes and one-hot entity category label from $j$-th ground truth relationships.

\subsection{Loss Calculation}
Our total loss $\mathcal{L}$ is composed of entity detector loss $\mathcal{L}^{enc}$ and predicate node generator loss $\mathcal{L}^{pre}$:
\begin{align}
    \mathcal{L} = \mathcal{L}^{enc} + \mathcal{L}^{pre}
\end{align}
The entity detector loss $\mathcal{L}^{enc}$ is calculated independently by following the same design in DETR\cite{carion2020end}.

The loss of predicate node generator $\mathcal{L}^{pre}$ is determined by the prediction and ground-the relationships according to the matching index $\mathbf{I}^{tri} \in \mathbb{N}^{N_{gt}}$.
The $\mathbf{I}^{tri}$ stores the index of matched predictions for each GT relationship.
Specifically, the predicate node generator loss consists of entity indicator loss $\mathcal{L}^{pre}_{i}$ and predicate sub-decoder loss $\mathcal{L}^{pre}_{p}$: 
\begin{align}
    \mathcal{L}^{pre}&=\mathcal{L}^{pre}_{i}+\mathcal{L}^{pre}_{p} 
\end{align}
For loss of predicate sub-decoder loss $\mathcal{L}^{pre}_{p}$, we have: 
\begin{align}
    \mathcal{L}^{pre}_{p} =  \sum_{i}^{N_{gt}} \left( \Vert \mathbf{b}_{p,\mathbf{I}^{tri}(i)}  - \mathbf{b}_{p,i}^{gt} \Vert_1  + \text{CE} \left( \mathbf{p}_{p,\mathbf{I}^{tri}(i)}, \mathbf{p}_{p,i}^{gt} \right)  \right)
\end{align}
where $\mathbf{b}_{p,i}^{gt}$ and $\mathbf{b}_{p,\mathbf{I}^{tri}(i)}$ is the entity center coordinates of the GT relationship and prediction, respectively. 
The $\text{CE}$ denotes the cross entropy loss between the predicate classification $\mathbf{b}_{p,\mathbf{I}^{tri}(i)}$ and the GT predicate category one-hot vector $\mathbf{p}_{p,i}^{gt}$.

For the entity indicator loss $\mathcal{L}^{pre}_{i}$:
\begin{align}
    \mathcal{L}^{pre}_{i} = \sum_{\star=\{s, o\}} \left( \mathcal{L}^{\star}_{ent\_loc} + \mathcal{L}^{\star}_{ent\_cls} \right) 
\end{align}
where $\star=\{s, o\}$ indicates the subject/object role of an entity in relationships.
The indicator loss is composed of two factors, $\mathcal{L}^{\star}_{ent\_loc}$ and $\mathcal{L}^{\star}_{ent\_cls}$, for two types of semantic representation: bounding boxes $\mathbf{b}_{s}, \mathbf{b}_{o}$ and classification $\mathbf{p}_{s}, \mathbf{p}_{o}$.
\begin{align}
    \mathcal{L}^{\star}_{ent\_loc} =  \sum_{i}^{N_{gt}} & \left( \Vert \mathbf{b}_{\star,\mathbf{I}^{tri}(i)}  - \mathbf{b}_{\star,i}^{gt} \Vert_1  \right.\\
    &\left. +1-~\text{GIOU}\left( \mathbf{b}_{\star,\mathbf{I}^{tri}(i)}, \mathbf{b}_{\star,i}^{gt} \right) \right) \\
    \mathcal{L}^{\star}_{ent\_cls} = \sum_{i}^{N_{gt}} & \text{CE}  \left( \mathbf{p}_{\star,\mathbf{I}^{tri}(i)}, \mathbf{p}_{\star,i}^{gt} \right)
\end{align}
The $\mathcal{L}^{\star}_{ent\_loc}$ is computed by the $L1$ distance and GIoU loss between the bounding box outputs $\mathbf{b}_{\star,\mathbf{I}^{tri}(i)}$ and ground-truth entity boxes $\mathbf{b}_{\star,i}^{gt}$.
The $\mathcal{L}^{\star}_{ent\_cls}$ is calculated from the cross entropy loss of classification prediction $\mathbf{p}_{\star,\mathbf{I}^{tri}(i)}$ according to ground-truth entities'category $\mathbf{p}_{\star,i}^{gt}$.

\section{Social Impacts}
Our method has no direct potential negative impact, one possible negative impact is that SGG may serve as a base module for surveillance abuse.